\documentclass[journal]{IEEEtran}
\usepackage{amsmath,amsfonts}
\usepackage{algorithm}
\usepackage{array}
\usepackage{textcomp}
\usepackage{stfloats}
\usepackage{url}
\usepackage{verbatim}
\usepackage{graphicx}
\usepackage{cite}
\usepackage{booktabs}
\usepackage{multirow}
\usepackage{xcolor}
\usepackage{makecell}
\usepackage{tabularx}
\usepackage{float}
\usepackage{placeins}
\usepackage{colortbl}
\usepackage[export]{adjustbox}
\usepackage{hyperref}
\usepackage{algorithm}
\usepackage{algpseudocode}
\usepackage{amsmath}
\usepackage{pdfpages}



\hypersetup{
    colorlinks=true,
    linkcolor=red,      
    citecolor=green,     
    filecolor=black,   
    urlcolor=black        
}
\begin{document}

\title{FAST-GOAL: Fast and Efficient Global-local Object Alignment Learning}
\author{Hyungyu Choi\textsuperscript{*}, Young Kyun Jang\textsuperscript{*}, Chanho Eom\textsuperscript{$\dagger$} \\
\thanks{\textsuperscript{*}Authors contributed equally. \hspace{1em} \textsuperscript{$\dagger$}Corresponding author. \\
Hyungyu Choi and Chanho Eom are with the Department of Virtual Convergence, Graduate School of Advanced Imaging Science, Multimedia \& Films (GSAIM), Chung-Ang University, Seoul, South Korea (e-mail: qkenr0804@cau.ac.kr; cheom@cau.ac.kr). Young Kyun Jang (e-mail: kyun0914@gmail.com). This paper has supplementary downloadable material available at http://ieeexplore.ieee.org., provided by the author. The material includes additional experiments related to this work. Contact qkenr0804@cau.ac.kr for further questions about this work. \\
GitHub: \url{https://github.com/PerceptualAI-Lab/FAST-GOAL}}}

\markboth{Journal of \LaTeX\ Class Files,~Vol.~14, No.~8, August~2021}%
{Shell \MakeLowercase{\textit{Choi et al.}}: FAST-GOAL: Fast and Efficient Global-local Object Alignment
Learning}

\maketitle

\begin{abstract}
Vision-language models such as CLIP have shown impressive capabilities in aligning images and text, but they often struggle with lengthy and detailed text descriptions due to pre-training on short and concise captions. We present FAST-GOAL (Fast and Efficient Global-local Object Alignment Learning), an efficient fine-tuning method that enhances ability of CLIP to handle lengthy text through global-local semantic alignment. Our method consists of two key components. First, Fast Local Image-Sentence Matching (FLISM) efficiently extracts local image regions through object detection and spatial division, then matches them with corresponding sentences. Second, Token Similarity-based Learning (TSL) maximizes the similarity between patch tokens from specific regions in the image and their corresponding region embeddings, applying the same principle to text, which enhances the ability of the model to capture detailed correspondences. Additionally, we introduce GLIT100k, a dataset that provides both global image-lengthy caption pairs and context-derived local pairs, where local descriptions are extracted from global captions to maintain semantic coherence. Through extensive experiments on long caption datasets (DOCCI, DCI) and short caption datasets (MSCOCO, Flickr30k), we demonstrate that FAST-GOAL achieves significant improvements over baselines, enabling effective adaptation of CLIP to detailed textual descriptions while maintaining computational efficiency.
\end{abstract}


\section{Introduction}
\label{sec:Introduction}

 \IEEEPARstart{A}{fter} the emergence of CLIP~\cite{CLIP}, numerous methods~\cite{ALBEF, LiT, Llip,  SigLIP} have been proposed to find the connection between images and text, showcasing significant advancements. By aligning hundreds of millions of image-caption pairs through contrastive learning, CLIP successfully encodes images and text into a unified embedding space. The resulting distribution of image and text embeddings captures both visual and textual semantics, enabling zero-shot transfer to various downstream tasks, such as retrieval~\cite{DeViSE, DeViSE+, retrieval1, retrieval2} and classification~\cite{AlexNet, VGGNet, Inception, ResNet}, while achieving decent performance.

 \begin{figure}[ht!]
\raggedleft
\begin{minipage}{0.48\columnwidth}
    \centering
    \includegraphics[width=4.3cm]{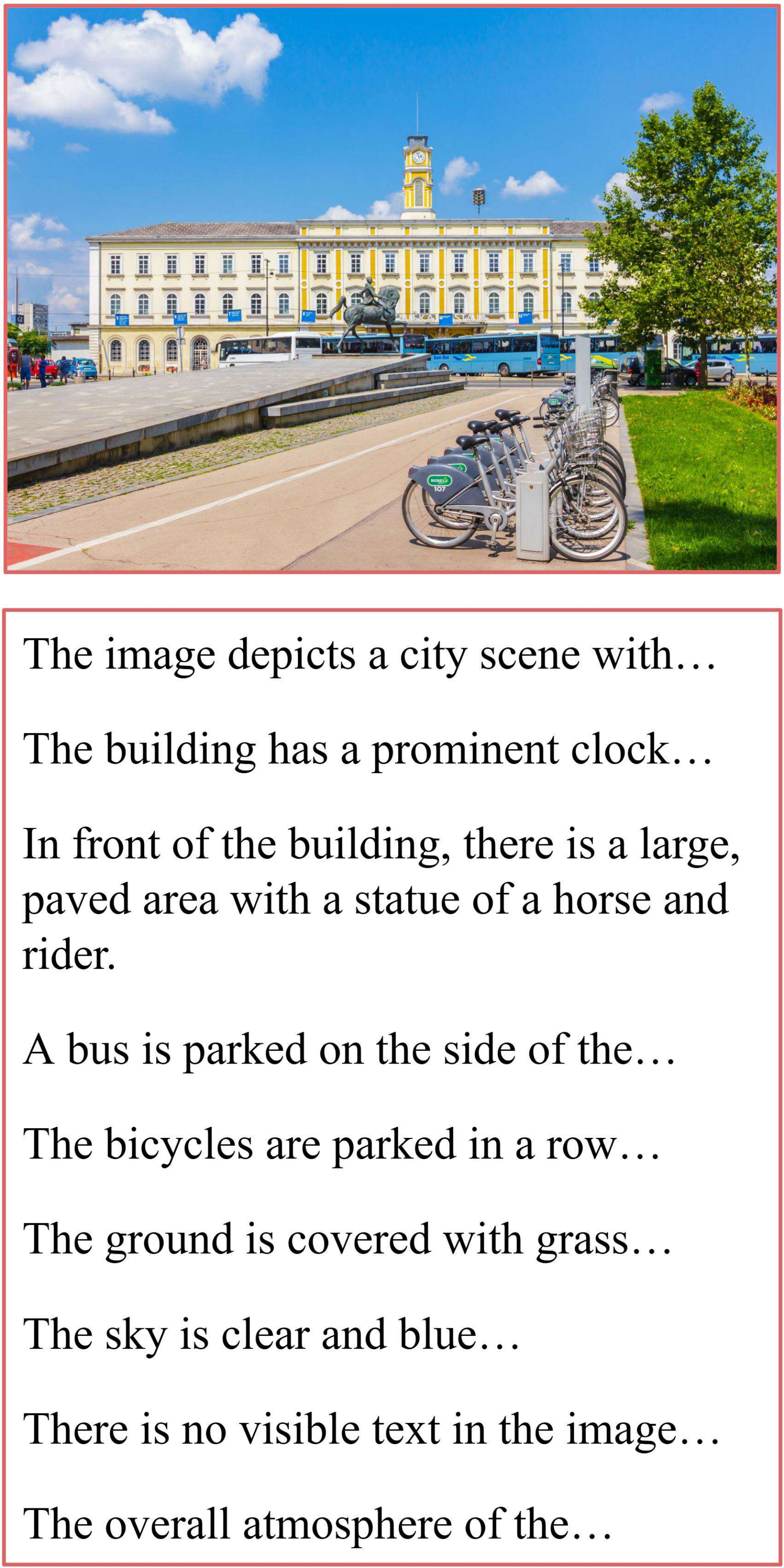}
    
    (a) CLIP
    \label{fig:intro1_a}
\end{minipage}
\hfill
\begin{minipage}{0.48\columnwidth}
    \centering
    \includegraphics[width=4.3cm]{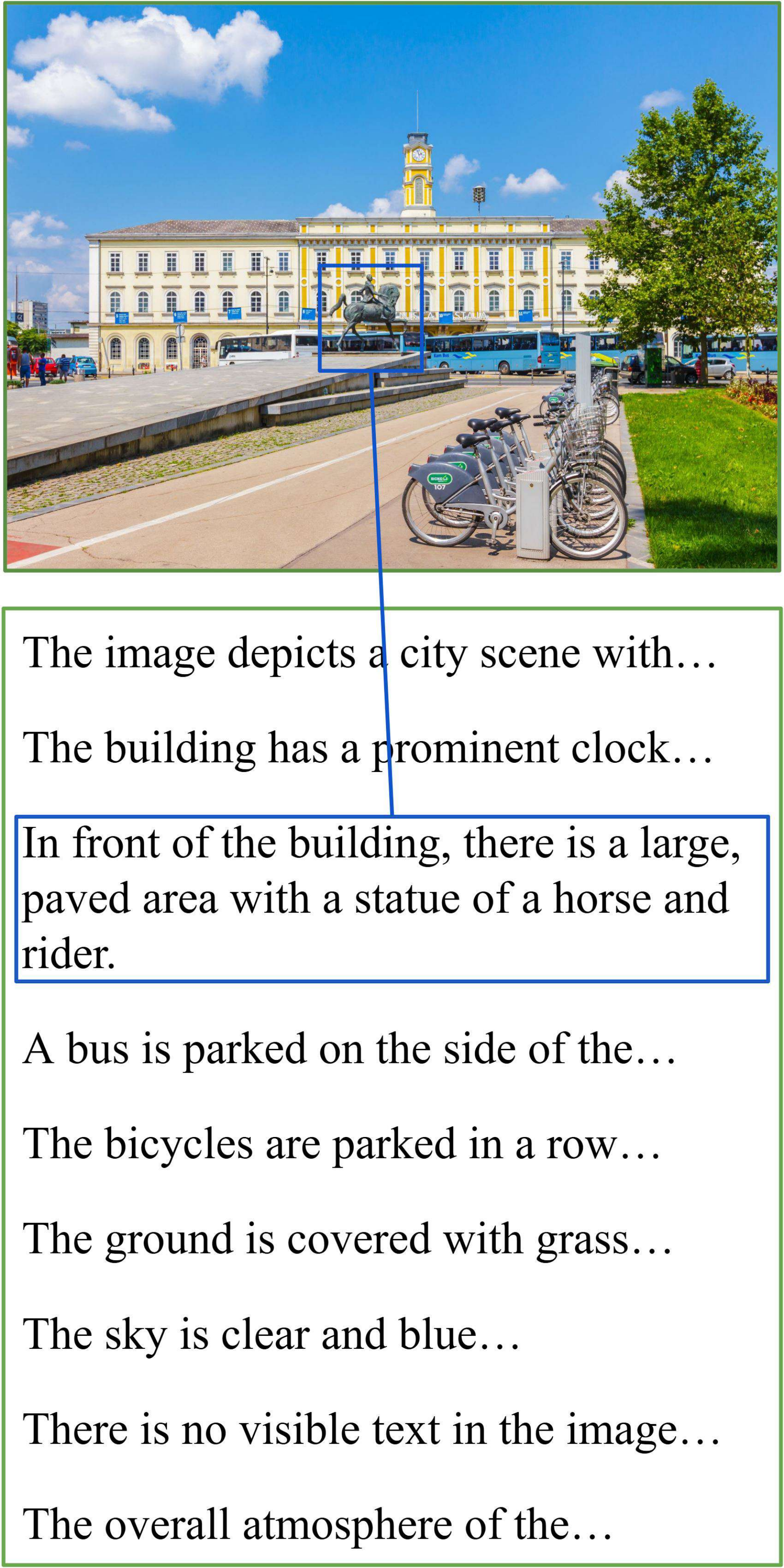}
    
    (b) FAST-GOAL
    \label{fig:intro_b}
\end{minipage}

\caption{Comparison of CLIP and our FAST-GOAL's capability in handling image-text alignment. (a) CLIP is limited to global image-text matching, treating the entire image and full caption as single units without detailed associations. (b) FAST-GOAL can establish precise local alignments between specific regions in the image and their corresponding textual descriptions in the caption (highlighted in blue).}
\label{fig:intro}
\end{figure}

However, fine-tuning a pre-trained CLIP (Fig.~\ref{fig:intro} (a)) model for specific domains faces limitations, as CLIP is trained on general, short captions~(\textit{e.g.}, maximum 77 tokens in the vanilla model) that focus on high-level image concepts. When tasked with longer, more detailed text, CLIP struggles to capture nuanced information, as the unified embedding space is optimized for concise descriptions. This makes adapting CLIP for retrieval tasks requiring lengthy text challenging without architectural adjustments or specialized training techniques.

In this paper, we propose a novel but simple fine-tuning method for image and lengthy text pairs, called FAST-GOAL (Fast and Efficient Global-local Object Alignment Learnig) (Fig.~\ref{fig:intro} (b)). Here, we refer to \textit{``global"} as the entire image or text and \textit{``local"} as a sub-part, such as a segment of the image or a specific sentence in the text. The idea behind FAST-GOAL is to enable the encoder model to focus on the dominant local elements within each image and text sample, thereby enhancing the overall understanding of the sample and producing a more representative embedding. 

FAST-GOAL has two key components: First, \textbf{F}ast \textbf{L}ocal \textbf{I}mage-\textbf{S}entence \textbf{M}atching (FLISM), a pipeline that extracts local segments from images and matches them with corresponding descriptive sentences from the entire caption. Second, we introduce \textbf{T}oken \textbf{S}imilarity-based \textbf{L}earning (TSL), a method that effectively propagates attention of local element using the local pairs obtained through the FLISM pipeline. Additionally, we present  \textbf{G}lobal-\textbf{L}ocal \textbf{I}mage-\textbf{T}ext pair \textbf{100k} (GLIT100k), a dataset comprising 100k image-lengthy caption pairs that also provides fine-grained local pairs matching image segments to corresponding sentences, enabling multi-level supervision for lengthy text understanding. To validate FAST-GOAL across diverse domains, we evaluate our method on both long caption datasets (DOCCI~\cite{DOCCI} and DCI~\cite{DCI}) and short caption datasets (COCO~\cite{COCO} and Flickr30k~\cite{Flickr30k}). This extensive evaluation demonstrates that FAST-GOAL effectively addresses the challenge of image-lengthy text retrieval while maintaining robust performance on short caption tasks. Our method shows substantial improvements compared to baseline models across diverse evaluation scenarios.
The main contributions of our work can be summarized as follows:
\begin{itemize}
\item We propose FAST-GOAL, a fast and efficient fine-tuning method that leverages FLISM and TSL methods to enable effective adaptation of CLIP for lengthy text understanding through global-local alignment at scale.
\item  We introduce GLIT100k, a dataset that provides both image-lengthy caption pairs and local pairs matching image segments to corresponding text descriptions, offering the multi-level supervision necessary for effective lengthy text understanding.
\item Through diverse experiments across both long caption domains including DOCCI and DCI, as well as short caption domains such as MSCOCO and Flickr30k, we show that FAST-GOAL significantly outperforms baseline CLIP models while maintaining strong performance across varying caption lengths.
\end{itemize}
This paper builds upon our prior work GOAL~\cite{GOAL}. This journal extension adds: 
1) FLISM, an efficient pipeline that combines object detection with spatial division for local pair matching, reducing computational costs while maintaining accurate image-text correspondences between regions and sentences;
2) GLIT100k, a 100k-scale dataset providing context-derived local pairs where local descriptions are extracted from global captions rather than generated independently, ensuring semantic coherence across global-local hierarchies;
3) Experimental validation across both long caption datasets (DOCCI, DCI) and short caption datasets (MSCOCO, Flickr30k), demonstrating consistent improvements across varying caption lengths;
4) Analysis of computational efficiency showing that our 100k-scale approach achieves competitive performance while requiring less resources than million-scale alternatives;
5) Qualitative analysis through retrieval visualizations demonstrating superior ability of FAST-GOAL to capture fine-grained visual-textual correspondences compared to baseline models.

\section{Related Work}
\label{sec:Related Work}

\subsection{Vision-Language Pre-training}
Research on addressing alignment differences between vision and language modalities has brought the CLIP~\cite{CLIP} model into the spotlight. CLIP, a multi-modal embedding model trained through contrastive learning on over 400 million image-text pairs, effectively aligns visual and textual representations while demonstrating remarkable zero-shot capabilities. Following its success, larger pre-training models emerged, such as ALIGN~\cite{ALIGN} and Florence~\cite{Florence}, trained on image-text pairs from datasets containing 1.8B and 900M samples, respectively. However, these models typically rely on short, broad image descriptions as captions, causing them to miss crucial local-level detailed information. This limits their ability to focus primarily on global understanding while failing to capture local details. To overcome this limitation, we present a fine-tuning method that enhances CLIP's ability to capture both local-detail and global-semantic information by training it on a dataset containing detailed, multi-sentence captions.

\subsection{CLIP for Long Text Understanding}
The primary challenge in enhancing CLIP~\cite{CLIP} for long text understanding lies in finding an effective balance between dataset scale and performance. Current methods fall into two main paradigms, each with fundamental trade-offs. Large scale pre-training methods such as FG-CLIP~\cite{FG-CLIP} first leverage massive billion-scale datasets to develop robust multi-modal representations from scratch, then require additional hard negative samples to enhance fine-grained understanding. This two-stage process results in demanding computational resources and training infrastructure, severely limiting practical adoption. As an alternative, fine-tuning methods have gained attention for their accessibility. Long-CLIP~\cite{LongCLIP} pioneers the use of lengthy captions generated by multimodal large language models (MLLMs) for CLIP fine-tuning, incorporating both coarse-grained and fine-grained alignment strategies with million-scale datasets~\cite{sharegpt4v}. FineLIP~\cite{FineLIP} follows a similar path, relying on million-scale data and employing additional token aggregation modules to reduce ambiguity before cross-modal alignment. Despite their improvements, the reliance on million-scale datasets in both methods raises fundamental questions about whether such large-scale data requirements are truly necessary for effective fine-tuning. Addressing this question, GOAL~\cite{GOAL} demonstrates that remarkable performance can be achieved using only small-scale datasets~\cite{DOCCI, DCI}, proving that data efficiency is possible without massive resources. Nevertheless, the expensive dataset curation process in GOAL creates a new bottleneck. This prevents exploration of the optimal balance point where both efficiency and scalability could coexist. To address this challenge, we propose FAST-GOAL, which leverages our 100k-scale dataset, GLIT100k. This scale is large enough to ensure robust performance yet small enough to maintain computational efficiency. Through efficient dataset construction and training processes, our method achieves superior performance without the burden of million-scale data or additional architectural parameters, establishing a new paradigm for practical long text understanding in CLIP.

\subsection{Utilizing Local Elements in Vision-Language Model Training} 
In terms of vision-language alignment models, using local elements' knowledge to improve the model's general ability has been widely explored across various domains. ViTAA~\cite{ViTAA} learns to align full-person images corresponding to the global-level with text describing the whole person to perform a person re-identification task~\cite{reid1, reid2, reid3, reid4}, while also learning to align the image and text for attributes~(\textit{e.g.},~hair, pants, shoes) that correspond to the local-level. This approach combines global-local relations, enabling richer visual-language representation learning. CLOC~\cite{CLOC} builds 2 billion image-text datasets and uses them for pre-training models by matching local objects and phrase-levels through Open-vocabulary Detector (\textit{e.g.}, OWLv2~\cite{OWLv2}, GLIPv2~\cite{GLIPv2}) models. This approach aims to improve localization capabilities while maintaining CLIP's global-level representation, demonstrating superior performance compared to the original pre-trained CLIP model. 
In contrast, our proposed FAST-GOAL method addresses image-lengthy text matching through region-sentence alignment, which fundamentally differs from existing fine-tuning approaches~\cite{FineCLIP, DenseVLM} that focus on region-phrase matching with short captions. While these methods align image regions with noun phrases or brief descriptions, lengthy captions contain multiple complete sentences describing various scene aspects. This requires sentence-level local pairs to maintain contextual coherence between global and local descriptions.

\section{Method}
In this section, we first present Fast Local Image-Sentence Matching (FLISM), an efficient pipeline that generates local-level pseudo pairs from image-lengthy caption pairs (Sec.~\ref{subsec:Fast Local Image-Sentence Matching}). We then describe Token Similarity-based Learning (TSL), which leverages these pseudo pairs to enhance CLIP's fine-grained understanding capabilities (Sec.~\ref{subsec:Token Similarity based Learning}).

\subsection{Fast Local Image Sentence Matching}
\label{subsec:Fast Local Image-Sentence Matching}

\begin{figure*}[!t]
\centering
\includegraphics[width=\textwidth]{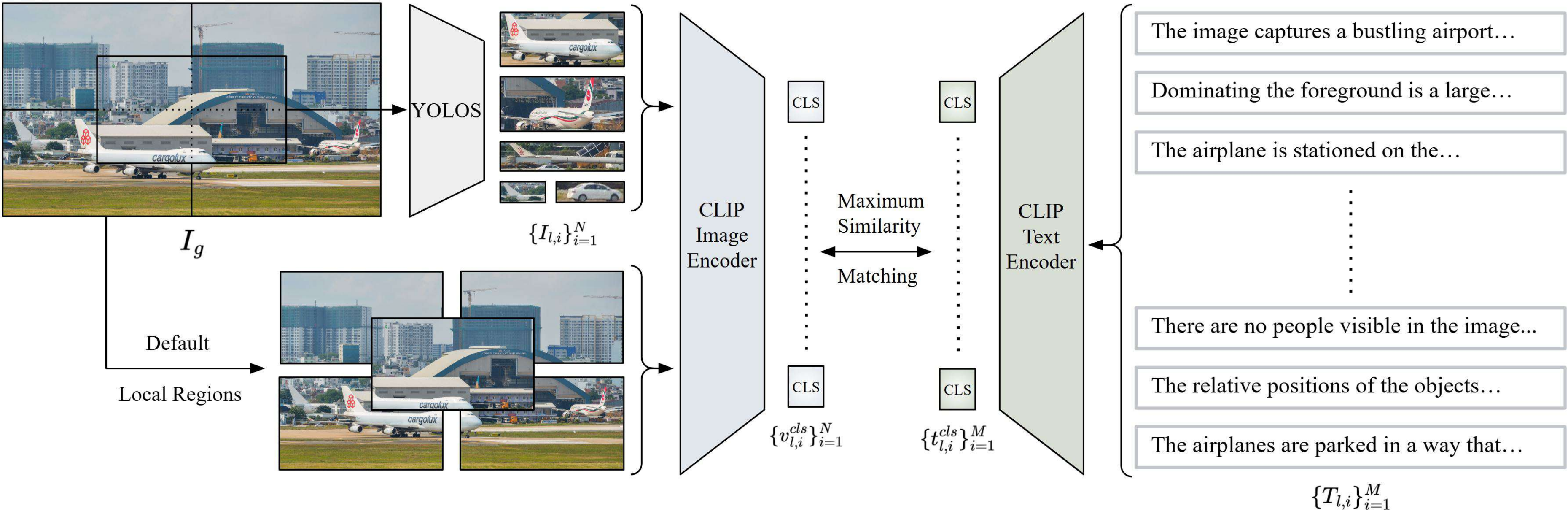}
\caption{Overview of Fast Local Image-Sentence Matching (FLISM) pipeline. Given a global image and its detailed caption, FLISM uses YOLOS~\cite{YOLOS} to detect objects in the image to create local regions and splits the caption into individual sentences. These local pairs are then processed through CLIP encoder to obtain CLS embeddings, which are used for maximum similarity matching to identify the most relevant image-sentence pairs.}
\label{fig:FLISM}
\end{figure*}

We propose Fast Local Image-Sentence Matching (FLISM) illustrated in Fig.~\ref{fig:FLISM}. FLISM efficiently separates a given caption into individual sentences and identifies corresponding image segments, matching each sentence with its relevant segment. To this end, we first decompose a given caption $T_g$, which provides detailed descriptions of a given image $I_g$, into individual sentences, resulting in text segments $\{T_{l,i}\}_{i=1}^M$, where $M$ is the number of sentences.

For local image extraction, FLISM combines object detection with spatial division to obtain diverse local image candidates. While the previous method~\cite{GOAL} relies on a dense segmentation methods, SAM~\cite{SAM}, that require significant computational overhead to analyze all possible image regions, we design an efficient extraction strategy that focuses on semantically meaningful regions. Specifically, we employ YOLOS~\cite{YOLOS}, a lightweight object detection model, to detect and extract object regions from the image $I_g$, directly identifying semantically meaningful local segments that correspond to actual visual entities. We complement these object-centric regions with default local regions: four quadrants dividing the image into equal sections and one central region formed by connecting the midpoints of each quadrant. The YOLOS-detected regions capture object-level information while the default regions provide spatial coverage for areas that may not contain distinct objects. Together, these yield a set of local images, $\{I_{l,i}\}_{i=1}^N$, where $N$ represents the total number of local regions. This approach enables local image extraction without the computational overhead of exhaustive segmentation methods.

We use CLIP~\cite{CLIP} to match the decomposed caption segments with the corresponding image segments. Specifically, we extract the CLS token embeddings for each local text segment $\{T_{l,i}\}_{i=1}^M$ from the text encoder of CLIP, $\phi_t$, as follows:
\begin{equation}
    \{t^{\text{cls}}_{l,i}\}_{i=1}^M = \phi_t(\{T_{l,i}\}_{i=1}^M).
\end{equation}
Similarly, for each image segment $\{I_{l,i}\}_{i=1}^N$, we extract the CLS token embeddings from the visual encoder of CLIP as follows:
\begin{equation}
    \{v^{\text{cls}}_{l,i}\}_{i=1}^N = \phi_v(\{I_{l,i}\}_{i=1}^N).
\end{equation}
Next, we compute the cosine similarity between each local text embedding $t^{\text{cls}}_{l,i}$ and the local image embeddings $\{v^{\text{cls}}_{l,i}\}_{i=1}^N$. Among all matched pairs, each local text embedding is matched with its highest similarity image embedding. From all these matched pairs, we select the one pair with the highest similarity score and denote it as $(I_l, T_l)$. This matching strategy ensures high-quality local pair associations by focusing exclusively on local correspondences, while maintaining computational efficiency through the YOLOS-based extraction process.

\subsection{Token Similarity based Learning}
\label{subsec:Token Similarity based Learning}

\begin{figure*}[!t]
\centering
\includegraphics[width=\textwidth]{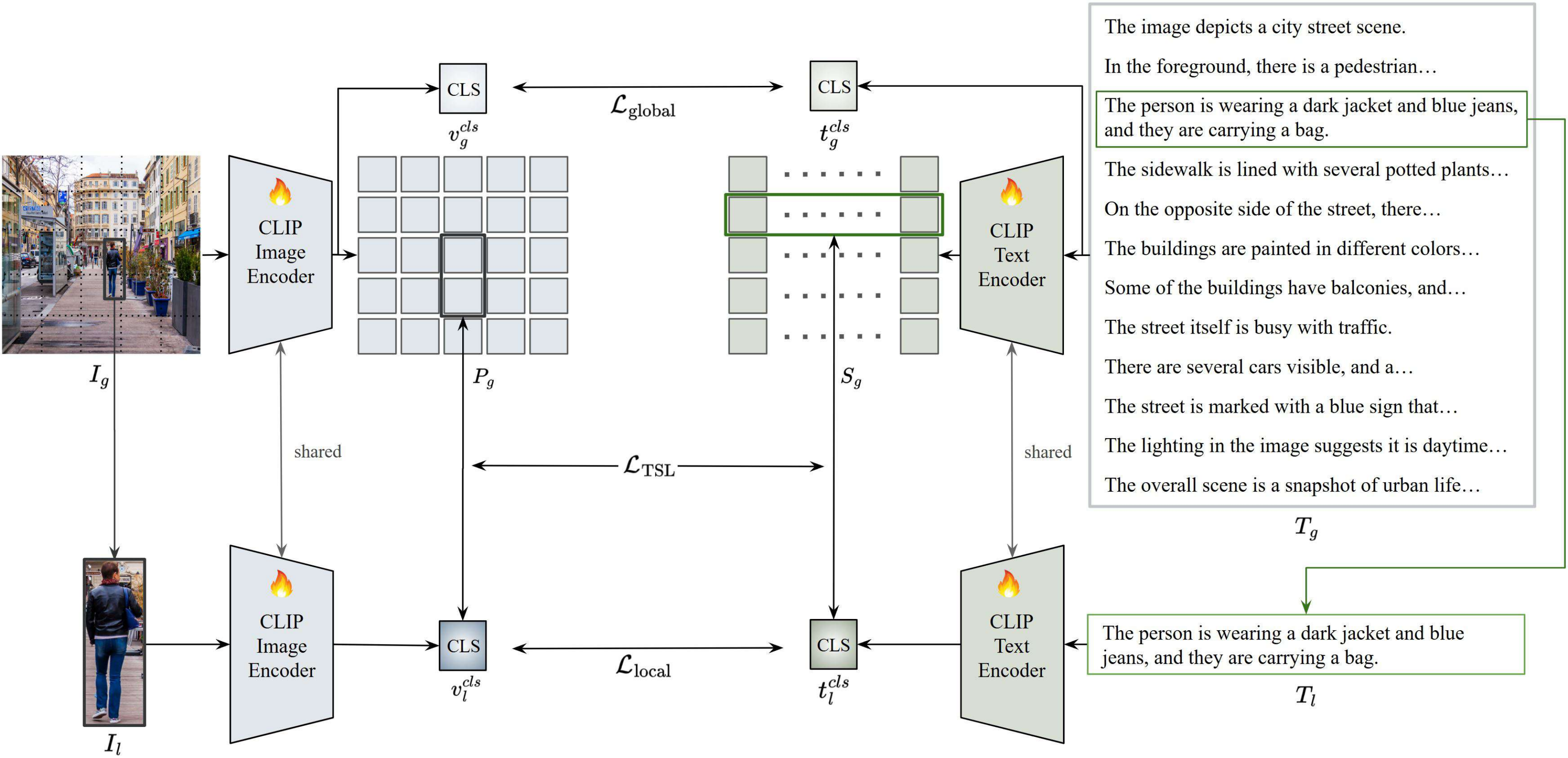}
\caption{Overview of Token Similarity based Learning (TSL). The framework processes global image-text pairs and their local pairs through shared CLIP encoders, extracting patch and sequence tokens. TSL identifies and projects corresponding token regions to match local CLS embeddings, enabling attention on local element.}
\label{fig:TSL}
\end{figure*}

While CLIP's pretraining with image-text pairs effectively learns global alignment, its training with brief captions limits the model's ability to capture fine-grained local details from lengthy descriptions. To address this, we propose Token Similarity based Learning (TSL) (Fig.~\ref{fig:TSL}). Our approach uses local pairs obtained through the FLISM pipeline and implements a fine-tuning strategy that effectively propagates local-level information. Specifically, TSL maximizes the similarity between patch tokens of local regions in the global image and their corresponding local image embeddings, while applying the same principle to text by increasing the similarity between sequence tokens of local parts in the global text and their corresponding local text embeddings.
To implement this strategy, we need to extract both local and global features from the input pairs.  Using CLIP's vision encoder $\phi_v$ and text encoder $\phi_t$, we extract both local and global features as follows:
For the local text $T_l$,
\begin{equation}
t^{cls}_l = \phi_t(T_l) \in \mathbb{R}^d,
\end{equation}
where $t^{cls}_l$ represents the last layer CLS token embedding. For the global text $T_g$, the text encoder extracts:
\begin{equation}
S_g = \phi_t(T_g) \in \mathbb{R}^{M \times d},
\end{equation}
where $M$ is the sequence length of $T_g$, and $S_g$ represents the last layer sequence tokens of $T_g$. To handle text sequences longer than CLIP's standard 77 token limit, we adopt Long-CLIP's~\cite{LongCLIP} positional embedding interpolation method in our text encoder.
For the local image $I_l$, we obtain:
\begin{equation}
v^{cls}_l = \phi_v(I_l) \in \mathbb{R}^d,
\end{equation}
where $v^{cls}_l$ represents the last layer CLS token embedding. For the global image $I_g$, the vision encoder extracts:
\begin{equation}
P_g = \phi_v(I_g) \in \mathbb{R}^{N \times d},
\end{equation}
where $N$ denotes the number of patch tokens in $I_g$ , $d$ is the embedding dimension and $P_g$ represents the last layer patch tokens of $I_g$. We process both global and local pairs through shared CLIP encoders to learn both types of features simultaneously. This weight sharing ensures consistent encoding in the shared embedding space. Let $\mathcal{T}$ denote the set of token indices corresponding to the local text segment. We can identify the sequence tokens in $S_g$ that correspond to $T_l$,
\begin{equation}
S_m = \frac{1}{|\mathcal{T}|}\sum_{i \in \mathcal{T}} S_g[i] \in \mathbb{R}^d,
\end{equation}
where $|\mathcal{T}|$ denotes the number of selected sequence tokens. The aggregated features are then projected into a shared embedding space, where both text and image representations are aligned: 
\begin{equation}
\hat{S_l} = \psi_t(S_m) \in \mathbb{R}^d,
\end{equation}
where $\psi_t(\cdot)$ represents a learned textual projection function.
 
Given that each local image region $I_l$ has its bounding box coordinates $(x_1, y_1, x_2, y_2)$ obtained from FLISM in the global image $I_g$, we can leverage this spatial information to identify specific patch tokens from $P_g$ that correspond to the local image region, filtering out patches from other parts of the global image. Let $\mathcal{B}$ denote the set of indices of patch tokens located inside the bounding box. We aggregate these tokens using average pooling to capture comprehensive information from the selected region:
\begin{equation}
P_m = \frac{1}{|\mathcal{B}|}\sum_{i \in \mathcal{B}} P_g[i] \in \mathbb{R}^d,
\end{equation}
where $|\mathcal{B}|$ denotes the number of selected patch tokens. The aggregated features are then projected into a shared embedding space where both text and image representations are aligned: 
\begin{equation}
\hat{P_l} = \psi_v(P_m ) \in \mathbb{R}^d,
\end{equation}
where the $\psi_v(\cdot)$ represents a learned visual projection function.
We train our model with multiple objectives combined into a final loss function:
\begin{equation}
\mathcal{L}_{\text{total}} =  \lambda_{global}\mathcal{L}_{\text{global}} + \lambda_{local}\mathcal{L}_{\text{local}} + \lambda_{TSL}\mathcal{L}_{\text{TSL}},
\end{equation}
where $\lambda$ is a hyperparameter controlling the contribution of local alignment.  We apply contrastive learning at both global and local levels, adopting the contrastive learning used in CLIP. At the global level:
\begin{equation}
\mathcal{L}_{\text{global}} = \mathcal{L}_{\text{contrast}}(v_{g}^{cls}, t_{g}^{cls}),
\end{equation}
where  $v_{g}^{cls}$ and  $t_{g}^{cls}$ are the CLS token embeddings of the global image $I_g$ and global text $T_g$, respectively. This global alignment ensures that the model maintains CLIP's original capability to capture global relationships between image-text pairs. Similarly, for local-level contrastive learning:
\begin{equation}
\mathcal{L}_{\text{local}} = \mathcal{L}_{\text{contrast}}(v_{l}^{cls},
t_{{l}}^{cls}),
\end{equation}
where  $v_{l}^{cls}$ and  $t_{l}^{cls}$ are the CLS token embeddings of the local image $I_l$ and local text $T_l$, respectively. By applying contrastive learning to local CLS token pairs, we encourage precise alignment between local image regions and their corresponding textual descriptions, enabling the model to learn cross-modal relationships.

\begin{figure}[t!] 
    \centering
    \includegraphics[width=0.82\linewidth]{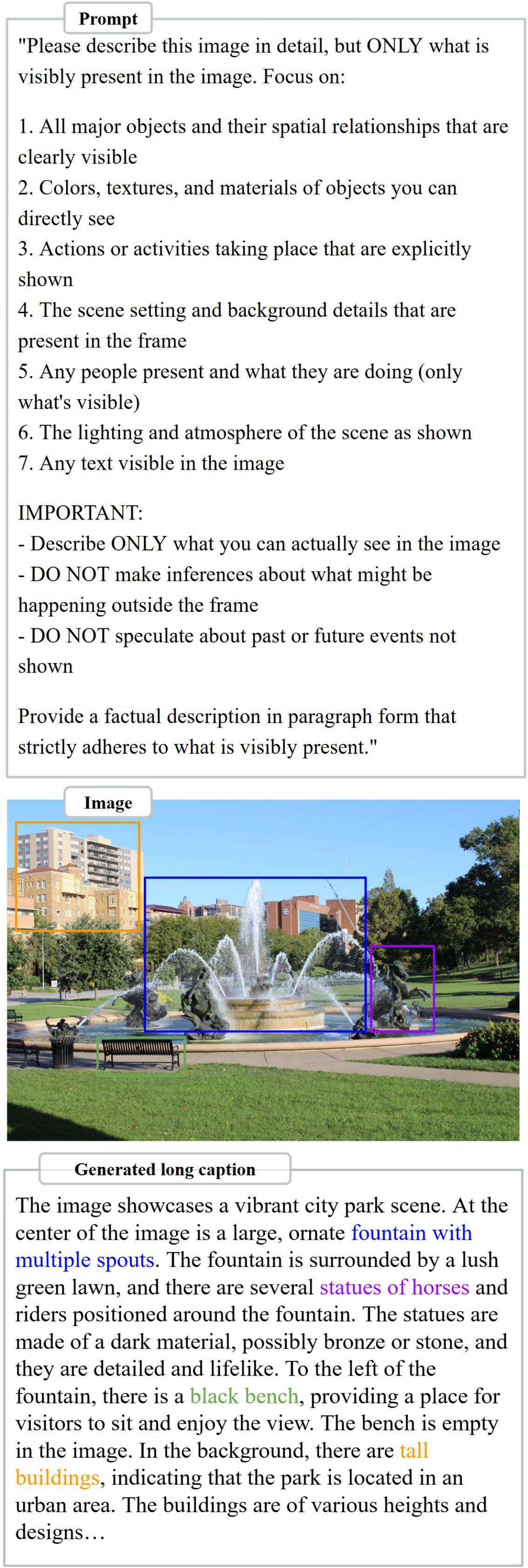} 
    \caption{Prompt-based caption generation with LLaVA-Next~\cite{LLaVA-NEXT}. Our prompt guides the model to generate descriptions covering scene context, object attributes, and spatial relationships while minimizing hallucinations.}
    \label{fig:promt}
\end{figure}

The token similarity loss $\mathcal{L}_{\text{TSL}}$ maximizes the similarity between projected tokens and their corresponding local CLS token embeddings for both image and text:
\begin{equation}
\mathcal{L}_{\text{TSL}} = \text{MSE}(\text{\textit{sim}}(\hat{P_l},v_{l}^{cls}), \mathbf{1}) + \text{MSE}(\text{\textit{sim}}(\hat{S_l},t_{{l}}^{cls}), \mathbf{1}),
\end{equation}
where $\text{\textit{sim}}(\cdot)$ denotes a function that computes an $n \times n$ similarity matrix with $n$ being the batch size, and $\mathbf{1}$ is a $n \times n$ matrix with ones on its diagonal entries. By optimizing this loss, the model learns to maximize the similarity between local CLS token embeddings and their corresponding regions in global tokens. This token-level alignment strategy enables the model to attention on local element, enhancing fine-grained understanding capabilities.
This fine-tuning method effectively addresses CLIP's inherent limitation in capturing local details from lengthy descriptions, which stems from its pre-training with brief captions. Through the combination of token-level similarity learning and global-local contrastive learning, our approach enables comprehensive understanding of cross-modal relationships with attention on local element from detailed text descriptions.

\section{GLIT100k Dataset Construction}
\label{sec:GLIT100k Dataset Construction}

\subsection{Long caption generation}
\label{subsec:Long caption generation}

We construct GLIT100k (Global-Local Image-Text pair 100k), a dataset that uniquely provides both image-lengthy caption pairs and fine-grained local pairs for enhancing long text understanding capabilities of CLIP~\cite{CLIP}. Our dataset construction derives local pairs directly from global captions, ensuring semantic consistency between different granularities while avoiding redundant generation processes. We sample diverse images from SA-1B~\cite{SAM}, which contains rich visual scenes with multiple objects and complex spatial arrangements. The diversity of SA-1B, which is originally curated for segmentation tasks across varied domains, provides an ideal foundation for extracting both global scene understanding and detailed local image-text correspondences. Recent advances in multi-modal large language models (MLLMs)~\cite{LLaVA, Qwen, InternVL, Gemini, LLaVA-NEXT} have enabled generation of highly detailed and accurate scene descriptions. We employ LLaVA-Next~\cite{LLaVA-NEXT} for caption generation, which has demonstrated strong performance in producing comprehensive visual descriptions. \par

Figure~\ref{fig:promt} illustrates our prompt that ensures high-quality lengthy captions by guiding the model to minimize hallucinations and produce visually-grounded descriptions. Our prompt explicitly instructs the model to describe only visually observable elements while covering multiple levels of detail from overall scene context to specific object attributes and spatial relationships. This structured approach yields detailed multi-sentence captions that capture comprehensive scene semantics, providing the rich textual information necessary for lengthy text understanding tasks.

\begin{table}[t!]
\centering
\caption{Caption quality comparison between generated and human-annotated datasets measured using Long-CLIP ViT-L/14 on 1,000 randomly sampled image-text pairs.}
\label{tab:caption_quality}

\begin{tabular}{lcc}
\toprule
Dataset & Annotation Type & Avg. Cosine Similarity \\
\midrule
GLIT100k & Generated (LLaVA-Next) & 0.2434 \\
DOCCI & Human-annotated & 0.2448 \\
\midrule
Difference & & 0.0014 \\
\bottomrule
\end{tabular}
\end{table}

To validate the quality of our generated captions, we conduct a comparison between GLIT100k and the human-annotated DOCCI~\cite{DOCCI} dataset. Table~\ref{tab:caption_quality} presents quantitative validation results measuring semantic alignment between images and their corresponding text descriptions. We randomly sample 1,000 image-text pairs from each dataset and compute their cosine similarity using Long-CLIP ViT-L/14. GLIT100k achieves 0.2434 average cosine similarity compared to the 0.2448 achieved by human-annotated DOCCI, demonstrating comparable semantic consistency with only 0.0014 difference. This minimal gap validates that our prompt engineering approach effectively produces high-quality captions that maintain strong semantic alignment with their corresponding images, ensuring reliable supervision for training vision-language models.

\subsection{Long caption datasets}
\label{subsec:Long Caption Dataset Analysis}

Table~\ref{tab:dataset_comparison} shows a comparison of datasets generated by MLLMs for long text understanding. Each dataset differs in scale and structural design. ShareGPT4V~\cite{sharegpt4v} contains 1.2M samples but only provides global caption pairs, lacking fine-grained image-text alignment. FineHARD~\cite{FG-CLIP} scales to 12M, including global and local pairs. However, local pairs are built by independently generating text for individual cropped image regions, resulting in potential semantic inconsistencies and loss of contextual relationships between local descriptions and the overall scene narrative. In contrast, our GLIT100k introduces context-derived local pairs and includes both global and local pairs. We utilize the specific local sentence within the broader narrative context of the full caption it belongs to. This context-aware approach results in a richer and more accurate semantic representation for that individual sentence. This design maintains the natural relationships present in the original scene description while avoiding redundant generation processes. By preserving contextual connections between global and local pairs, GLIT100k is designed to achieve and effective long text understanding through its efficient structure without requiring million-scale data, as demonstrated by our experimental results in Sec.\ref{sec:Ablation Study}.

\begin{table}[!t]
\centering
\caption{Comparison of MLLM-generated datasets for CLIP long text understanding. Context-derived local pairs indicate descriptions extracted from global captions rather than generated independently.}
\normalsize
\renewcommand{\arraystretch}{1.2} 
\setlength{\tabcolsep}{4pt} 

\begin{tabularx}{\columnwidth}{l|c|c|c|>{\centering\arraybackslash}X}
\toprule

\multicolumn{1}{c|}{Dataset} & 
\multicolumn{1}{c|}{Scale} & 
\multicolumn{1}{c|}{\makecell{Global \\ Pairs}} & 
\multicolumn{1}{c|}{\makecell{Local \\ Pairs}} & 
\multicolumn{1}{c}{\makecell{Context-Derived \\ Local Pairs}} \\
\midrule
ShareGPT4V & 1.2M & \checkmark & & \\
FineHARD & 12M & \checkmark & \checkmark & \\
\midrule
GLIT100k (Ours) & 100K & \checkmark & \checkmark & \checkmark \\
\bottomrule
\end{tabularx}
\vspace{-2mm}
\label{tab:dataset_comparison}
\end{table}

\section{Experiments}

In this section, we present our experimental setup in Sec.\ref{sec:Experimental setup}. We then evaluate the zero-shot retrieval performance of FAST-GOAL across both long and short caption domains in Sec.\ref{sec:Zero-shot Retrieval Performance}, demonstrating the effectiveness of our method on datasets with varying caption lengths. Our ablation study analyzes the computational efficiency of Fast Local Image-Sentence Matching (FLISM) and validates the effectiveness of our GLIT100k dataset for multi-stage training. Additionally, we examine the contribution of each loss component in our framework in Sec.\ref{sec:Ablation Study}. Finally, we provide qualitative analysis through attention map visualizations and retrieval result comparisons in Sec.\ref{sec:Qualitative Results}.
\subsection{Experimental setup}
\label{sec:Experimental setup}

\noindent {\bf{Dataset.}}
We train our FAST-GOAL model using GLIT100k, our proposed dataset comprising 100,647 image-lengthy caption pairs with an average of 159 words per caption, along with fine-grained local pairs matching image segments to corresponding sentences. This dataset enables robust training on detailed textual descriptions and fine-grained visual-textual relationships at scale. For evaluation, we conduct experiments on both long caption domains including DOCCI~\cite{DOCCI} and DCI~\cite{DCI}, as well as short caption domains such as MSCOCO~\cite{COCO} and Flickr30k~\cite{Flickr30k}, to comprehensively assess our model's performance across varying caption lengths.

\noindent {\bf{Training Setting.}}
To validate our approach, we conduct experiments using CLIP~\cite{CLIP} ViT-B/16~\cite{ViT} backbone architecture. The model is fine-tuned for 10 epochs with a batch size of 256. We set the balance hyperparameters in the total loss function as $\lambda_{\mathrm{global}}=1$, $\lambda_{\mathrm{TSL}}=1$, and $\lambda_{\mathrm{local}}=0.5$ to maintain strong global and TSL learning while moderating the contribution of local loss. Training was performed on four NVIDIA A6000 GPUs, taking approximately 1--2 hours to complete.

\noindent {\bf{Test Setting.}} 
To handle the long text sequences during inference, we adopt the positional embedding interpolation from Long-CLIP~\cite{LongCLIP}. Our evaluation encompasses both text-to-image (T2I) and image-to-text (I2T) retrieval tasks, measured using Recall@K metrics (K = 1, 5, 10, 15, 25, 50). To ensure fair comparison across all baseline methods, we employ the ViT-B/16~\cite{ViT} backbone throughout our experiments. We evaluate our method on the original test sets of each dataset. We use the DOCCI test set consisting of 5,000 images, 7,602 samples from the DCI dataset after excluding 203 samples without long captions from the original 7,805 samples as our test set, MSCOCO 2014 validation set, and the Flickr30k test set.

\subsection{Zero-Shot Retrieval Performance}
\label{sec:Zero-shot Retrieval Performance}
We evaluate FAST-GOAL's effectiveness through comprehensive zero-shot retrieval experiments across datasets with varying caption lengths. We compare FAST-GOAL against several state-of-the-art baselines including the original CLIP~\cite{CLIP}, EVA-CLIP~\cite{EVA-CLIP}, Long-CLIP~\cite{LongCLIP}, FineCLIP~\cite{FineCLIP}, and our previous work~\cite{GOAL}.

\begin{table*}[!t]
\centering
\caption{Performance comparison on Long Caption Domain datasets (DOCCI and DCI) using ViT-B/16 backbone. Results show Recall@K scores for text to image and image to text retrieval tasks. The best performance is shown in \textbf{bold}, and the second-best performance is \underline{underlined}.}
\normalsize
\renewcommand{\arraystretch}{0.95}
\begin{tabular}{c|c|c|cccc|cccc}
\toprule
\multirow{2}{*}{Dataset} & \multirow{2}{*}{Method} & \multirow{2}{*}{Backbone} & \multicolumn{4}{c|}{Text to Image Recall@K} & \multicolumn{4}{c}{Image to Text Recall@K} \\
\noalign{\vskip 1pt} 
\cline{4-11}
\noalign{\vskip 2pt} 
& & & R@1 & R@5 & R@25 & R@50 & R@1 & R@5 & R@25 & R@50 \\
\midrule
\multirow{6}{*}{DOCCI} 
& CLIP & \multirow{6}{*}{ViT-B/16} & 58.53 & 85.57 & 97.16 & 98.82 & 58.80 & 85.18 & 96.73 & 98.45 \\
& EVA-CLIP & & 66.61 & 89.08 & 97.53 & 99.00 & 66.24 & 89.43 & 97.69 & 98.88 \\
& Long-CLIP & & \underline{71.63} & 92.16 & \underline{98.90} & \textbf{99.73} & 63.29 & 88.80 & 98.39 & 99.45 \\
& FineCLIP & & 62.92 & 87.22 & 96.67 & 98.35 & 56.55 & 82.94 & 95.47 & 97.92 \\
& GOAL & & 71.22 & \underline{92.39} & \underline{98.90} & 99.61 & \underline{72.18} & \underline{92.88} & \underline{98.88} & \underline{99.55} \\
\cmidrule{2-2} \cmidrule{4-11}
& FAST-GOAL & & \textbf{74.27} & \textbf{93.76} & \textbf{99.27} & \underline{99.71} & \textbf{74.67} & \textbf{94.71} & \textbf{99.24} & \textbf{99.75} \\
\midrule
\multirow{6}{*}{DCI}
& CLIP & \multirow{6}{*}{ViT-B/16} & 33.36 & 53.87 & 72.34 & 79.33 & 35.07 & 54.28 & 71.84 & 78.91 \\
& EVA-CLIP & & 41.25 & 61.19 & 77.62 & 83.03 & 40.71 & 61.14 & 77.73 & 83.60 \\
& Long-CLIP & & 48.41 & 68.71 & 82.49 & 87.16 & 42.21 & 67.21 & 83.21 & 88.21 \\
& FineCLIP & & 37.70 & 58.30 & 75.82 & 81.60 & 36.48 & 58.20 & 76.14 & 82.74 \\
& GOAL & & \underline{49.13} & \underline{70.17} & \underline{84.53} & \underline{88.69} & \underline{51.87} & \underline{71.10} & \underline{84.78} & \underline{89.12} \\
\cmidrule{2-2} \cmidrule{4-11}
& FAST-GOAL & & \textbf{56.62} & \textbf{75.26} & \textbf{87.75} & \textbf{91.08} & \textbf{56.74} & \textbf{75.99} & \textbf{88.71} & \textbf{92.46} \\
\bottomrule
\end{tabular}
\vspace{-2mm}

\vspace{-2mm}
\label{tab:long_caption_performance}
\end{table*}

\begin{table*}[!t]
\centering
\caption{Performance comparison on Short Caption Domain datasets (MSCOCO and Flickr30k) using ViT-B/16 backbone. Results show Recall@K scores for text to image and image to text retrieval tasks. The best performance is shown in \textbf{bold}, and the second-best performance is \underline{underlined}. Note that FineCLIP on MSCOCO is fine-tuned on MSCOCO itself, thus it is not a fair zero-shot comparison.}
\label{tab:short_caption_performance}
\normalsize
\renewcommand{\arraystretch}{0.95}
\begin{tabular}{c|c|c|cccc|cccc}
\toprule
\multirow{2}{*}{Dataset} & \multirow{2}{*}{Method} & \multirow{2}{*}{Backbone} & \multicolumn{4}{c|}{Text to Image Recall@K} & \multicolumn{4}{c}{Image to Text Recall@K} \\
\noalign{\vskip 1pt} 
\cline{4-11}
\noalign{\vskip 2pt} 
& & & R@1 & R@5 & R@10 & R@15 & R@1 & R@5 & R@10 & R@15 \\
\midrule
\multirow{5}{*}{MSCOCO}
& CLIP & \multirow{6}{*}{ViT-B/16} & 33.95 & 59.46 & 69.93 & 76.00 & 54.14 & 77.74 & 85.60 & 89.36 \\
& EVA-CLIP & & \underline{42.58} & \underline{67.49} & \underline{76.91} & \underline{81.81} & \underline{58.44} & \underline{81.06} & \underline{88.28} & \underline{91.66} \\
& Long-CLIP & & 40.83 & 66.36 & 76.22 & 81.51 & 57.24 & 80.42 & 87.54 & 90.96 \\
& GOAL & & 38.86 & 64.36 & 74.15 & 79.58 & 59.28 & 81.02 & 87.70 & 90.88 \\
\cmidrule{2-2}\cmidrule{4-11}
& FAST-GOAL & & \textbf{42.81} & \textbf{68.61} & \textbf{78.10} & \textbf{82.92} & \textbf{61.00} & \textbf{82.84} & \textbf{89.24} & \textbf{92.34} \\
\midrule
\multirow{6}{*}{Flickr30k}
& CLIP & \multirow{6}{*}{ViT-B/16} & 63.20 & 86.30 & 92.38 & 94.38 & 82.90 & 97.20 & 98.90 & \underline{99.40} \\
& EVA-CLIP & & \underline{71.82} & \underline{91.38} & \textbf{95.48} & \underline{96.78} & \textbf{87.80} & \textbf{98.30} & \underline{99.20} & \underline{99.40} \\
& Long-CLIP & & 70.80 & 90.68 & 94.74 & 96.58 & 85.90 & 98.50 & \textbf{99.50} & \textbf{99.80} \\
& FineCLIP & & 71.16 & \textbf{91.42} & \underline{95.40} & \textbf{96.92} & 84.60 & 97.00 & 98.90 & \underline{99.40} \\
& GOAL & & 68.32 & 89.30 & 93.74 & 95.62 & 85.10 & 96.70 & 98.60 & 99.10 \\
\cmidrule{2-2}\cmidrule{4-11}
& FAST-GOAL & & \textbf{72.50} & 91.34 & 95.02 & 96.70 & \underline{86.80} & \underline{97.50} & 98.90 & \textbf{99.80} \\
\bottomrule

\end{tabular}
\end{table*}

\noindent {\bf{Long Caption Datasets.}}
Table~\ref{tab:long_caption_performance} presents results on datasets with lengthy and detailed descriptions: DOCCI~\cite{DOCCI} and DCI~\cite{DCI}. These datasets are particularly challenging as they contain comprehensive visual descriptions that require fine-grained understanding of image-text correspondences. For fair a evaluation of our previous work~\cite{GOAL}, we employ cross-dataset evaluation where models trained on DOCCI are evaluated on DCI and vice versa. The results demonstrate the superior performance of FAST-GOAL in handling lengthy textual descriptions, significantly outperforming baseline methods across all recall metrics. On the challenging R@1 and R@5 metrics, which require precise matching, FAST-GOAL demonstrates significant performance gains compared to other baseline methods. On DOCCI, FAST-GOAL achieves 74.27\% and 93.76\% for text-to-image R@1 and R@5, significantly outperforming the 71.63\% and 92.16\% of Long-CLIP respectively. Similarly, on DCI, FAST-GOAL attains 56.62\% and 75.26\% compared to the 48.41\% and 68.71\% of Long-CLIP. These substantial gaps in the more difficult recall metrics demonstrate that our method excels particularly in scenarios requiring precise retrieval accuracy. Notably, FAST-GOAL shows substantial improvements over our previous work~\cite{GOAL}, validating the effectiveness of our training approach with the GLIT100k dataset and enhanced FLISM pipeline.

\

\noindent {\bf{Short Caption Datasets.}} Table~\ref{tab:short_caption_performance} shows results on traditional short caption datasets: MSCOCO~\cite{COCO} and Flickr30k~\cite{Flickr30k}. These datasets contain concise descriptions typical of conventional image-text retrieval tasks. Despite being optimized for lengthy text understanding, FAST-GOAL maintains competitive performance on short caption domains, demonstrating the generalizability of our approach across varying caption lengths. On MSCOCO, FAST-GOAL achieves 42.81\% and 68.61\% for text-to-image R@1 and R@5, outperforming EVA-CLIP's 42.58\% and 67.49\% respectively (we exclude FineCLIP from this comparison as it is fine-tuned on MSCOCO itself, which would not constitute a fair zero-shot evaluation). On Flickr30k, FAST-GOAL attains 72.50\% for text-to-image R@1, outperforming EVA-CLIP's 71.82\%. The results indicate that our method does not sacrifice performance on traditional short caption retrieval tasks while gaining substantial improvements on lengthy text understanding. This balanced performance consistency makes FAST-GOAL a practical solution for real-world applications where both short and long text descriptions may be encountered. The consistent performance across varying text lengths demonstrates that FAST-GOAL effectively adapts to different levels of textual detail without overfitting to specific caption styles.

\begin{table*}[!t]
\centering
\caption{Ablation study comparing LISM and FLISM processing efficiency and evaluating the effectiveness of GLIT100k fine-tuning on DOCCI and DCI datasets using ViT-B/16 backbone. FAST-GOAL$^{\dagger}$ denotes direct fine-tuning on target dataset, while FAST-GOAL$^{\dagger\dagger}$ denotes fine-tuning on GLIT100k first, followed by additional fine-tuning on target dataset. The best performance is shown in \textbf{bold}, and the second-best performance is \underline{underline}.}
\normalsize
\renewcommand{\arraystretch}{0.95} 
\begin{tabular}{c|c|c|cccc|cccc}
\toprule
\multirow{2}{*}{Methods} & \multirow{2}{*}{\begin{tabular}[c]{@{}c@{}} Preprocessing\\Method\end{tabular}} & \multirow{2}{*}{\begin{tabular}[c]{@{}c@{}}Preprocessing\\Time (min)\end{tabular}} & \multicolumn{4}{c|}{Text to Image Recall@K} & \multicolumn{4}{c}{Image to Text Recall@K} \\
\cmidrule(lr){4-7} \cmidrule(lr){8-11}
& & & R@1 & R@5 & R@25 & R@50 & R@1 & R@5 & R@25 & R@50 \\
\midrule
\multicolumn{11}{c}{\textit{DOCCI}} \\
\midrule
GOAL & LISM & 1239.7 & 81.12 & \underline{97.12} & \underline{99.71} & 99.88 & \underline{81.00} & \underline{97.39} & \underline{99.75} & \underline{99.94} \\
FAST-GOAL$^{\dagger}$ & FLISM & 97.63 & \underline{81.65} & \underline{97.12} & \textbf{99.76} & \textbf{99.96} & 80.71 & 97.20 & \underline{99.75} & \textbf{99.98} \\
FAST-GOAL$^{\dagger\dagger}$ & FLISM & 97.63 & \textbf{84.16} & \textbf{97.69} & \textbf{99.76} & \underline{99.92} & \textbf{83.16} & \textbf{97.49} & \textbf{99.78} & \underline{99.94} \\
\midrule
\multicolumn{11}{c}{\textit{DCI}} \\
\midrule
GOAL & LISM & 407.71 & \underline{73.19} & 89.14 & 95.50 & \underline{97.25} & \underline{73.04} & \underline{90.10} & \underline{96.80} & \underline{98.15} \\
FAST-GOAL$^{\dagger}$ & FLISM & 34.44 & 72.99 & \underline{89.64} & \underline{95.55} & 97.20 & 72.44 & 89.69 & 96.35 & 98.10 \\
FAST-GOAL$^{\dagger\dagger}$ & FLISM & 34.44 & \textbf{78.29} & \textbf{91.00} & \textbf{96.50} & \textbf{97.80} & \textbf{77.34} & \textbf{91.05} & \textbf{97.55} & \textbf{98.60} \\
\bottomrule
\end{tabular}
\vspace{-2mm}
\label{tab:training_strategy}
\end{table*}

\subsection{Ablation Study}
\label{sec:Ablation Study}
\begin{table*}[!t]
\centering
\caption{Loss function ablation study on DOCCI and MSCOCO datasets using ViT-B/16 backbone. Results demonstrate the effectiveness of combining Global, Local, and TSL loss components across different text complexity domains. The best performance is shown in \textbf{bold}, and the second-best performance is \underline{underlined}.}
\normalsize
\renewcommand{\arraystretch}{0.95} 
\begin{tabular}{c|ccc|cccc|cccc}
\toprule
\multirow{2}{*}{Methods} & \multicolumn{3}{c|}{Loss Components} & \multicolumn{4}{c|}{Text to Image Recall@K} & \multicolumn{4}{c}{Image to Text Recall@K} \\
\cmidrule(lr){2-4} \cmidrule(lr){5-8} \cmidrule(lr){9-12}
& Global & Local & TSL & R@1 & R@5 & R@25 & R@50 & R@1 & R@5 & R@25 & R@50 \\
\midrule
\multicolumn{12}{c}{\textit{DOCCI (Long Caption Domain)}} \\
\midrule
Global fine-tuning & \checkmark & & & 76.02 & 95.25 & 99.47 & 99.84 & 75.51 & 94.61 & 99.29 & 99.76 \\
Local fine-tuning & & \checkmark & & 70.33 & 92.73 & 99.00 & 99.70 & 71.67 & 92.76 & 99.00 & 99.71 \\
w/o TSL & \checkmark & \checkmark & & \underline{78.27} & \underline{96.27} & \textbf{99.73} & \underline{99.92} & \underline{78.37} & \underline{96.20} & \textbf{99.61} & \underline{99.84} \\
FAST-GOAL & \checkmark & \checkmark & \checkmark & \textbf{79.06} & \textbf{96.31} & \underline{99.65} & \textbf{99.94} & \textbf{78.94} & \textbf{96.35} & \underline{99.57} & \textbf{99.88} \\
\midrule
\multicolumn{12}{c}{\textit{MSCOCO (Short Caption Domain)}} \\
\midrule
Global fine-tuning & \checkmark & & & 37.39 & 62.70 & 85.44 & 92.73 & 55.66 & 79.02 & 93.60 & 97.06 \\
Local fine-tuning & & \checkmark & & 37.91 & 63.25 & 85.66 & 92.76 & 56.64 & 79.82 & 94.50 & 97.76 \\
w/o TSL & \checkmark & \checkmark & & \underline{39.04} & \underline{64.46} & \underline{86.56} & \underline{93.47} & \underline{57.30} & \underline{80.94} & \textbf{95.38} & \textbf{98.10} \\
FAST-GOAL & \checkmark & \checkmark & \checkmark & \textbf{39.38} & \textbf{64.75} & \textbf{86.75} & \textbf{93.71} & \textbf{59.62} & \textbf{82.18} & \underline{94.64} & \underline{97.86} \\
\bottomrule
\end{tabular}
\vspace{-2mm}
\label{tab:loss_ablation}
\end{table*}

\noindent {\bf{FLISM and GLIT100k.}} Table~\ref{tab:training_strategy} presents our ablation studies examining both the computational efficiency of our proposed FLISM compared to the LISM from GOAL~\cite{GOAL}, and the effectiveness of our GLIT100k dataset for training. We evaluate three different approaches: GOAL using the LISM with fine-tuning on the target dataset, FAST-GOAL$^{\dagger}$ using our FLISM with direct fine-tuning only on the target dataset, and FAST-GOAL$^{\dagger\dagger}$ using our FLISM with two-stage fine-tuning where we first fine-tune on our GLIT100k dataset followed by further fine-tuning on the target dataset. For experimental settings, we conduct experiments on both DOCCI~\cite{DOCCI} and DCI~\cite{DCI} datasets. We randomly sample 2k images as the test set for DCI, and train all models with a batch size of 64 for fair comparison. Note that all approaches are trained and evaluated on the same target dataset, representing non-cross-dataset evaluation where models are fine-tuned on the specific domain they are evaluated on.

Our experimental results demonstrate both the computational efficiency of FLISM and the effectiveness of our GLIT100k dataset. The processing time analysis reveals remarkable efficiency gains when comparing the LISM of GOAL with the FLISM of FAST-GOAL$^{\dagger}$ across both datasets. Specifically, the preprocessing time measures the duration required to process the target dataset. On the DOCCI dataset, FAST-GOAL$^{\dagger}$ requires only 97.63 minutes compared to the 1239.7 minutes required by GOAL, representing a speedup of approximately 12.7×. Similarly, for DCI dataset processing, FAST-GOAL$^{\dagger}$ achieves substantial efficiency improvements, requiring only 34.44 minutes compared to the 407.71 minutes required by GOAL, representing a speedup of approximately 11.8×. Despite these dramatic time reductions, FAST-GOAL$^{\dagger}$ maintains comparable performance to GOAL, demonstrating that our FLISM can achieve significant computational efficiency without sacrificing performance quality compared to the LISM. This comparable performance validates that YOLO~\cite{YOLOS}-based object detection effectively identifies semantically important regions for local alignment without requiring SAM's~\cite{SAM} category-agnostic segmentation capability.

Building on this efficiency foundation, the comparison between FAST-GOAL$^{\dagger}$ and FAST-GOAL$^{\dagger\dagger}$ reveals the substantial benefits of fine-tuning on our GLIT100k dataset followed by further fine-tuning on target datasets. FAST-GOAL$^{\dagger}$ achieves comparable performance to GOAL with significantly less processing time through direct target dataset fine-tuning. More impressively, FAST-GOAL$^{\dagger\dagger}$ demonstrates even greater advantages by incorporating an additional training stage on GLIT100k. Through this two-stage training approach, FAST-GOAL$^{\dagger\dagger}$ achieves significantly superior performance while still maintaining the computational efficiency of FLISM. Specifically, FAST-GOAL$^{\dagger\dagger}$ achieves 84.16\% text-to-image R@1 compared to the 81.65\% achieved by FAST-GOAL$^{\dagger}$ and the 81.12\% achieved by GOAL, while maintaining the same 12.7× speedup over preprocessing time of GOAL. On DCI, FAST-GOAL$^{\dagger\dagger}$ attains 78.29\% compared to the 72.99\% achieved by FAST-GOAL$^{\dagger}$ and the 73.19\% achieved by GOAL while maintaining the same 11.8× speedup over GOAL's LISM method. These consistent improvements across different datasets confirm that our GLIT100k enables models to learn more generalizable fine-grained image-text alignments. The key to this effectiveness lies in GLIT100k's unique combination of global and context-derived local pairs, which preserves natural relationships between local descriptions and overall scene narratives. This enhanced learning effectively transfers to various long caption domains.

\noindent {\bf{Loss Function.}}
We conduct a ablation study to analyze the contribution of each loss component in our FAST-GOAL framework. Table~\ref{tab:loss_ablation} compares different combinations of our proposed loss functions: Global loss, Local loss, and Token Similarity-based Learning (TSL) loss, evaluated on both long caption (DOCCI)~\cite{DOCCI} and short caption (MSCOCO)~\cite{COCO} domains using ViT-B/16~\cite{ViT} backbone. For DOCCI evaluation, we train and test models on the same dataset, while MSCOCO results represent zero-shot evaluation performance.

 The results demonstrate that our complete FAST-GOAL method, utilizing all three loss components, achieves the best performance across both domains. Interestingly, we observe domain-specific behaviors that validate our approach design. On the DOCCI dataset, which contains lengthy and detailed descriptions, the global fine-tuning method shows stronger baseline performance compared to local fine-tuning, achieving 76.02\% versus 70.33\% on text-to-image R@1 and 75.51\% versus 71.67\% on image-to-text R@1. This indicates that comprehensive textual understanding benefits from global semantic alignment. Conversely, on MSCOCO with its shorter captions compared to DOCCI, local fine-tuning demonstrates relatively better performance with 37.91\% versus 37.39\% on text-to-image R@1 and 56.64\% versus 55.66\% on image-to-text R@1, indicating that local alignment correspondences are more critical when dealing with concise descriptions.

\begin{table*}[h]
\centering
\caption{Ablation study on spatial partitioning strategy using ViT-B/16 backbone. Results demonstrate that combining YOLO detections with spatial partitions consistently improves performance across all datasets and metrics. The best performance is shown in \textbf{bold}.}
\label{tab:spatial_partition_ablation}
\normalsize
\renewcommand{\arraystretch}{0.95}

\begin{tabular}{c|c|cccc|cccc}
\toprule
\multirow{2}{*}{Dataset} & \multirow{2}{*}{Region Extraction} & \multicolumn{4}{c|}{Text to Image Recall@K} & \multicolumn{4}{c}{Image to Text Recall@K} \\
\noalign{\vskip 1pt} 
\cline{3-10}
\noalign{\vskip 2pt} 
& & R@1 & R@5 & R@25 & R@50 & R@1 & R@5 & R@25 & R@50 \\
\midrule
\multirow{2}{*}{DOCCI} 
& YOLO only & 69.41 & 91.39 & 98.47 & 99.45 & 71.88 & 92.82 & 98.73 & 99.61 \\
& YOLO + Spatial & \textbf{74.27} & \textbf{93.76} & \textbf{99.27} & \textbf{99.71} & \textbf{74.67} & \textbf{94.71} & \textbf{99.24} & \textbf{99.75} \\
\midrule
\multirow{2}{*}{DCI}
& YOLO only & 53.21 & 73.40 & 85.99 & 89.79 & 53.91 & 73.90 & 87.33 & 91.29 \\
& YOLO + Spatial & \textbf{56.62} & \textbf{75.26} & \textbf{87.75} & \textbf{91.08} & \textbf{56.74} & \textbf{75.99} & \textbf{88.71} & \textbf{92.46} \\
\midrule
\multirow{2}{*}{MSCOCO}
& YOLO only & 39.89 & 65.52 & 87.08 & 93.62 & 58.04 & 80.44 & 94.58 & 98.18 \\
& YOLO + Spatial & \textbf{42.81} & \textbf{68.61} & \textbf{88.54} & \textbf{94.64} & \textbf{61.00} & \textbf{82.84} & \textbf{95.34} & \textbf{98.32} \\
\midrule
\multirow{2}{*}{Flickr30k}
& YOLO only & 69.22 & 89.44 & 96.76 & 98.00 & 85.30 & 96.80 & 99.90 & 99.90 \\
& YOLO + Spatial & \textbf{72.50} & \textbf{91.34} & \textbf{97.92} & \textbf{98.80} & \textbf{86.80} & \textbf{97.50} & \textbf{100.00} & \textbf{100.00} \\
\bottomrule
\end{tabular}
\end{table*}
\begin{table*}[t!]
\centering
\caption{Ablation study comparing single highest-similarity pair versus top-3 pairs matching strategies using ViT-B/16 backbone. Top-3 (uniform) applies equal weight to all three pairs, while Top-3 (weighted) applies cosine similarity-based weights. The best performance is shown in \textbf{bold}.}
\label{tab:topk_matching_ablation}
\normalsize
\renewcommand{\arraystretch}{0.95}

\begin{tabular}{c|c|cccc|cccc}
\toprule
\multirow{2}{*}{Dataset} & \multirow{2}{*}{Matching Strategy} & \multicolumn{4}{c|}{Text to Image Recall@K} & \multicolumn{4}{c}{Image to Text Recall@K} \\
\noalign{\vskip 1pt} 
\cline{3-10}
\noalign{\vskip 2pt} 
& & R@1 & R@5 & R@25 & R@50 & R@1 & R@5 & R@25 & R@50 \\
\midrule
\multirow{3}{*}{DOCCI} 
& Top-3 (uniform) & 71.51 & 92.57 & 99.02 & 99.71 & 73.04 & 93.86 & 99.10 & 99.75 \\
& Top-3 (weighted) & 71.75 & 92.49 & 99.00 & 99.71 & 73.04 & 93.82 & 99.18 & 99.75 \\
& Single pair (FAST-GOAL) & \textbf{74.27} & \textbf{93.76} & \textbf{99.27} & \textbf{99.71} & \textbf{74.67} & \textbf{94.71} & \textbf{99.24} & \textbf{99.75} \\
\midrule
\multirow{3}{*}{DCI}
& Top-3 (uniform) & 55.06 & 74.34 & 87.21 & 90.96 & 54.72 & 74.74 & 88.53 & 92.45 \\
& Top-3 (weighted) & 55.25 & 74.43 & 87.15 & 90.96 & 54.87 & 74.77 & 88.56 & 92.41 \\
& Single pair (FAST-GOAL) & \textbf{56.62} & \textbf{75.26} & \textbf{87.75} & \textbf{91.08} & \textbf{56.74} & \textbf{75.99} & \textbf{88.71} & \textbf{92.46} \\
\midrule
\multirow{3}{*}{MSCOCO}
& Top-3 (uniform) & 41.18 & 67.00 & 87.38 & 93.93 & 57.18 & 80.08 & 94.70 & 97.62 \\
& Top-3 (weighted) & 41.24 & 66.99 & 87.38 & 93.94 & 57.52 & 80.06 & 94.66 & 97.66 \\
& Single pair (FAST-GOAL) & \textbf{42.81} & \textbf{68.61} & \textbf{88.54} & \textbf{94.64} & \textbf{61.00} & \textbf{82.84} & \textbf{95.34} & \textbf{98.32} \\
\midrule
\multirow{3}{*}{Flickr30k}
& Top-3 (uniform) & 70.14 & 90.40 & 97.36 & 98.56 & 84.50 & 97.40 & 99.90 & \textbf{100.00} \\
& Top-3 (weighted) & 70.60 & 90.32 & 97.40 & 98.58 & 84.80 & 97.10 & 99.90 & \textbf{100.00} \\
& Single pair (FAST-GOAL) & \textbf{72.50} & \textbf{91.34} & \textbf{97.92} & \textbf{98.80} & \textbf{86.80} & \textbf{97.50} & \textbf{100.00} & \textbf{100.00} \\
\bottomrule
\end{tabular}
\end{table*}

 Most importantly, the combination of global and local losses with our proposed TSL component consistently outperforms individual loss functions across both long and short caption domains. On DOCCI, FAST-GOAL achieves 79.06\% text-to-image R@1 and 78.94\% image-to-text R@1, showing improvements over the w/o TSL baseline of 78.27\% and 78.37\% respectively. Similarly, on MSCOCO, FAST-GOAL demonstrates significant gains with 39.38\% text-to-image R@1 and 59.62\% image-to-text R@1 compared to 39.04\% and 57.30\% without TSL. This validates our hypothesis that effective multi-modal learning requires both global semantic understanding and fine-grained local correspondences. TSL serves as a mechanism for propagating attention signals from local elements. The consistent improvements demonstrate the generalizability of our approach across varying caption lengths.

\noindent {\bf{Spatial Partitioning.}} To validate the contribution of spatial partitioning in our FLISM pipeline, we conduct ablation experiments comparing YOLO~\cite{YOLOS}-only extraction against our combined approach that incorporates both YOLO detections and spatial partitions (quadrant and center regions). Table~\ref{tab:spatial_partition_ablation} presents comprehensive results across all evaluation datasets using ViT-B/16 backbone.

The results demonstrate that spatial partitioning provides consistent and substantial improvements across all datasets and metrics. On long caption datasets, adding spatial partitions to YOLO detections yields significant performance gains. Specifically, on DOCCI, the combined approach achieves 74.27\% text-to-image R@1 compared to the 69.41\% achieved by YOLO-only extraction, representing a substantial improvement. Similarly, on DCI, our method attains 56.62\% compared to the 53.21\% achieved with YOLO detections alone. These improvements extend to short caption datasets as well, with MSCOCO achieving 42.81\% text-to-image R@1 compared to 39.89\% from YOLO-only approach, and Flickr30k attaining 72.50\% compared to 69.22\%. The consistent performance gains across both long and short caption domains validate the effectiveness of our combined extraction strategy.

These consistent improvements across diverse datasets demonstrate that spatial partitions capture semantically relevant information that effectively complements object-centric regions detected by YOLO. The heuristic partitions serve two critical functions: First, they ensure comprehensive spatial coverage of the image, capturing regions that may not contain distinct objects but are nonetheless described in the captions. Second, they help capture contextual relationships between objects and their spatial arrangements, which are often crucial for understanding detailed scene descriptions. Rather than introducing noise, these spatial partitions provide valuable supervision signals that enhance the model's ability to learn fine-grained global-local alignments. The consistent performance gains across both long and short caption domains validate the effectiveness of our combined extraction strategy.

\noindent {\bf{Matching Strategy.}}
To validate our design choice of selecting the single highest-similarity pair in FLISM, we conduct experiments comparing it against Top-k matching strategies. Table~\ref{tab:topk_matching_ablation} presents comprehensive results across all evaluation datasets using ViT-B/16 backbone. We implement and evaluate two top-3 matching strategies: uniform weighting where all three pairs contribute equally to the TSL loss during training, and similarity-proportional weighting where each pair's contribution is scaled by its matching confidence score.

\begin{figure}[!t]
\centering
\includegraphics[width=0.45\textwidth]{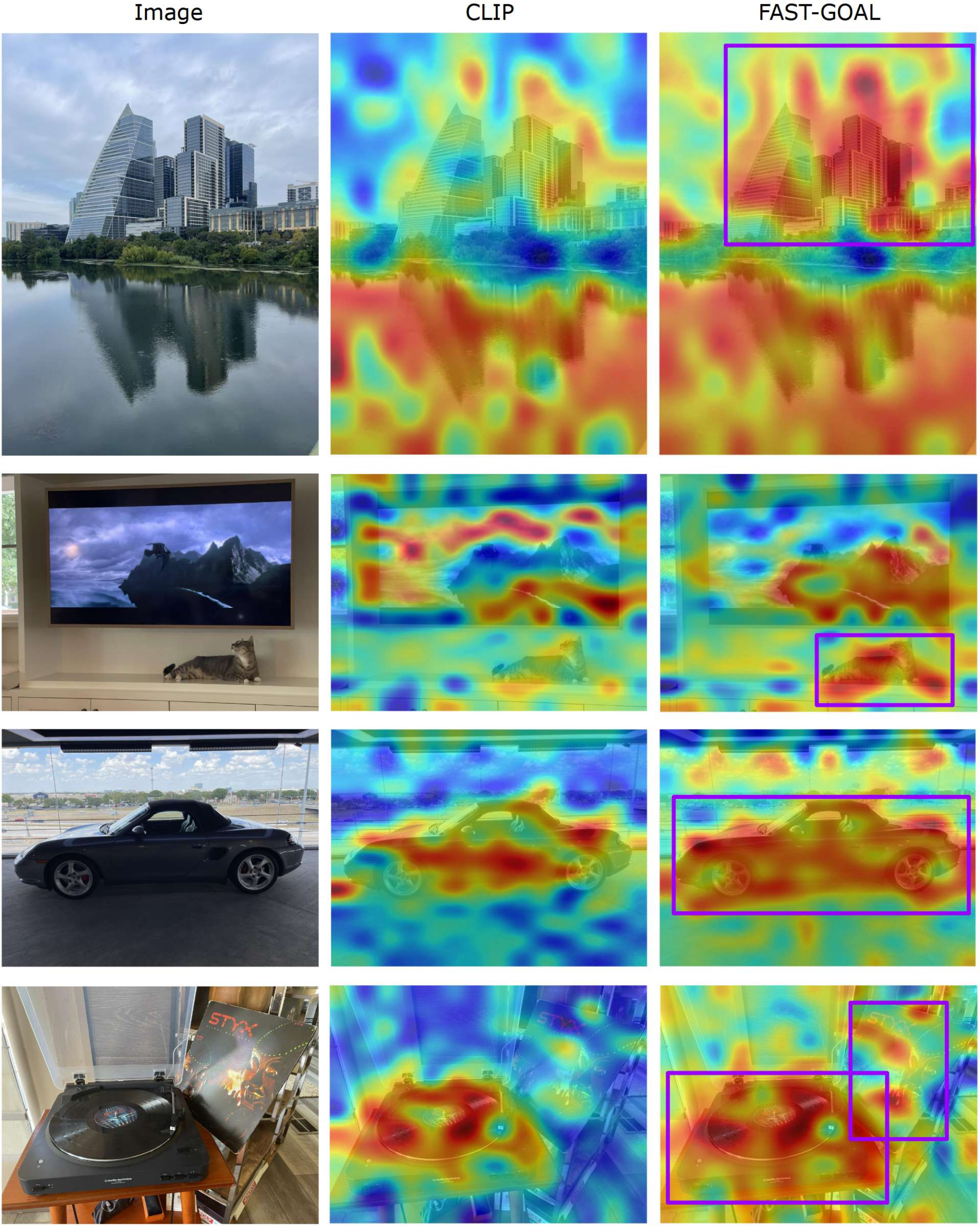}
\caption{Image encoder attention map visualization comparison between original CLIP and FAST-GOAL. Each row shows the input image, CLIP attention map, and FAST-GOAL attention map. Brighter regions indicate stronger attention weights within the image encoder's self-attention mechanism.}
\label{fig:attention_visualization}
\vspace{-2mm}
\end{figure}

The results consistently demonstrate that single pair selection outperforms both Top-3 strategies across all datasets and metrics. On DOCCI, single pair achieves 74.27\% text-to-image R@1 compared to the 71.75\% achieved by weighted matching and the 71.51\% achieved by uniform matching. On DCI, single pair attains 56.62\% compared to the 55.25\% from weighted matching and the 55.06\% from uniform matching. Similar patterns are observed on MSCOCO, where single pair achieves 42.81\% compared to 41.24\% and 41.18\% from the top-3 approaches, and on Flickr30k with 72.50\% compared to 70.60\% and 70.14\%.

These consistent improvements across diverse datasets validate that the highest-similarity pair effectively captures the most semantically relevant correspondence between image regions and text segments. Our single pair selection strategy focuses training on the strongest matches, enabling the model to learn from clear and unambiguous region-sentence associations. This focused approach proves more effective than distributing attention across multiple pairs, where varying similarity scores can affect the quality of supervision signals. Beyond performance advantages, single pair selection offers superior computational efficiency by computing TSL loss for only one pair instead of three, reducing memory consumption and training time while achieving better accuracy. This validates single pair selection as the optimal design choice for our FLISM pipeline.

\subsection{Qualitative Results}
\label{sec:Qualitative Results}

\noindent {\bf{Attention Map Visualization.}} Figure~\ref{fig:attention_visualization} presents side-by-side comparisons of how the image encoder attends to different regions within the same image. To understand how FAST-GOAL affects the internal attention mechanisms of the image encoder, we visualize and compare attention maps from the original CLIP~\cite{CLIP} model and our FAST-GOAL. We analyze the image encoder's self-attention patterns using the top 3 principal components~\cite{PCA} to capture the most significant attention variations across image patches.

 The attention visualizations show differences in visual processing. In the first row, CLIP produces weak attention across the entire image without focusing on specific buildings. In contrast, FAST-GOAL concentrates attention on the building and skyscrapers (purple box). In the second row, CLIP's attention scatters across the room without recognizing either the TV content or the cat below. FAST-GOAL, however, primarily attends to the mountain landscape on the screen while also capturing the cat (purple box). These distinct attention patterns reveal how our global-local alignment fundamentally changes the model's visual processing capabilities.

\begin{figure*}[!t]
\centering
\includegraphics[width=0.65\textwidth]{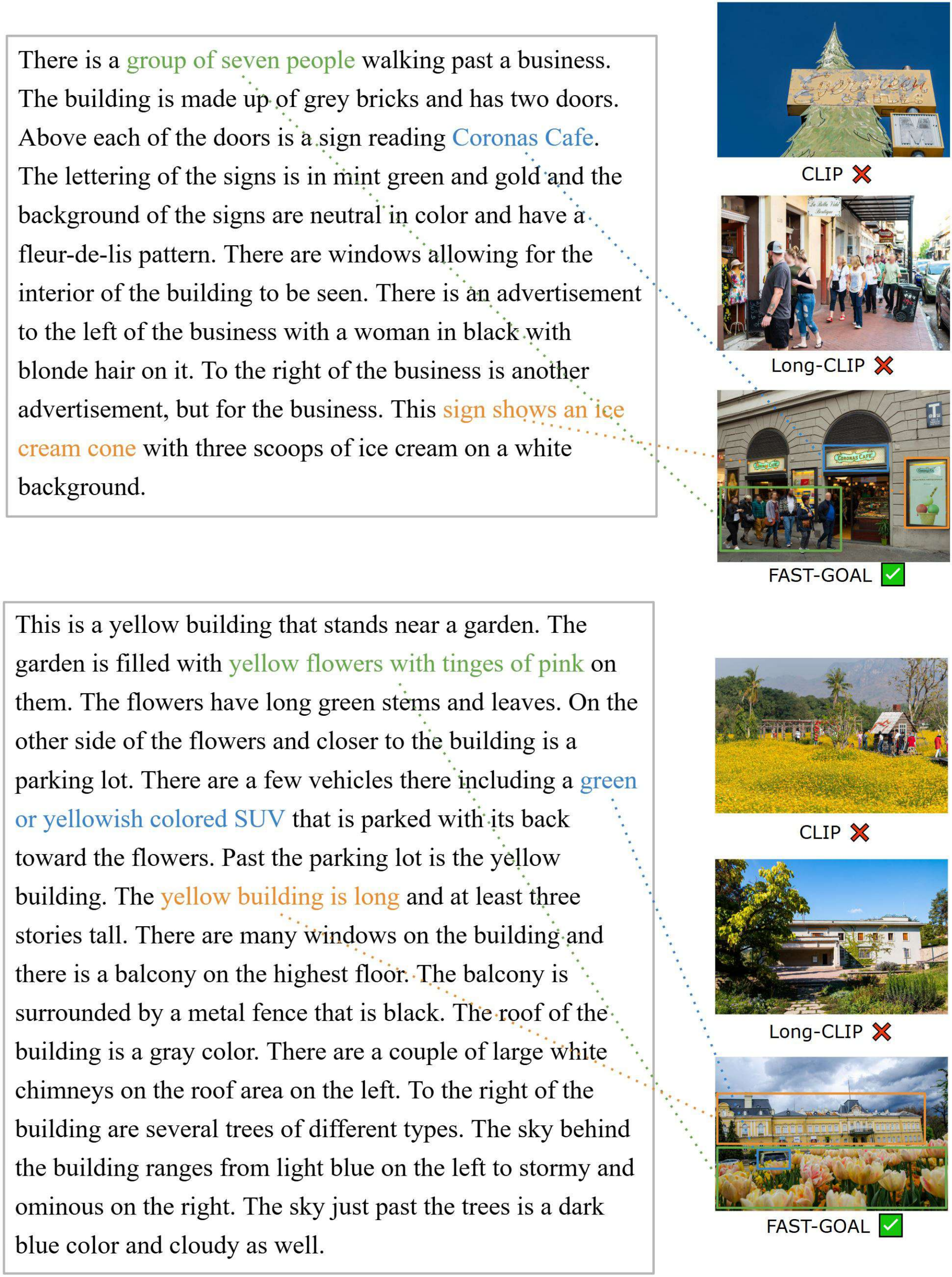}
\caption{Qualitative comparison of text-to-image retrieval results. Given lengthy text queries, we show the top-1 retrieved images from CLIP, Long-CLIP, and FAST-GOAL. The ground truth images demonstrate the target visual content that should be retrieved. Red boxes indicate incorrect retrievals, while green boxes show correct matches.}
\label{fig:qualitative_retrieval}
\vspace{-2mm}
\end{figure*}

The enhanced attention patterns in FAST-GOAL indicate that our method has learned to allocate stronger focus on visually salient elements that are more likely to be described in detailed textual descriptions. This enhanced attention behavior stems from our global-local alignment training strategy. Through training with our GLIT100k dataset and the TSL method, the image encoder learns to develop more discriminative internal representations that emphasize visually salient elements. This training process naturally guides the model to focus effectively on regions that are most relevant for comprehensive visual understanding. These attention visualizations demonstrate that FAST-GOAL not only improves text-image matching performance but also fundamentally enhances the ability of the image encoder to identify and emphasize semantically relevant visual content, leading to more robust visual representations specifically optimized for lengthy text understanding tasks.

\noindent {\bf{Retrieval Results.}} To provide intuitive insights into the effectiveness of FAST-GOAL, we present qualitative comparisons of text-to-image retrieval results using lengthy text queries. Figure~\ref{fig:qualitative_retrieval} shows representative examples where we compare the top-1 retrieved images from CLIP~\cite{CLIP}, Long-CLIP~\cite{LongCLIP}, and our FAST-GOAL method. The qualitative results reveal important insights about the capabilities of different methods when handling lengthy and detailed text descriptions.

CLIP and Long-CLIP demonstrate reasonable semantic understanding by retrieving images that contain some relevant visual elements mentioned in the query. However, they often fail to capture the fine-grained details described in the lengthy captions. For example, in the first query describing a group of seven people walking past a business with specific details about Coronas Cafe signage and grey brick building, we observe notable differences in retrieval accuracy. CLIP retrieves a completely irrelevant image, while Long-CLIP finds a similar image with people walking past a business but fails to capture the specific context and precise details such as the exact number of people and the distinctive cafe signage.

\begin{figure*}[!t]
\centering
\includegraphics[width=\linewidth]{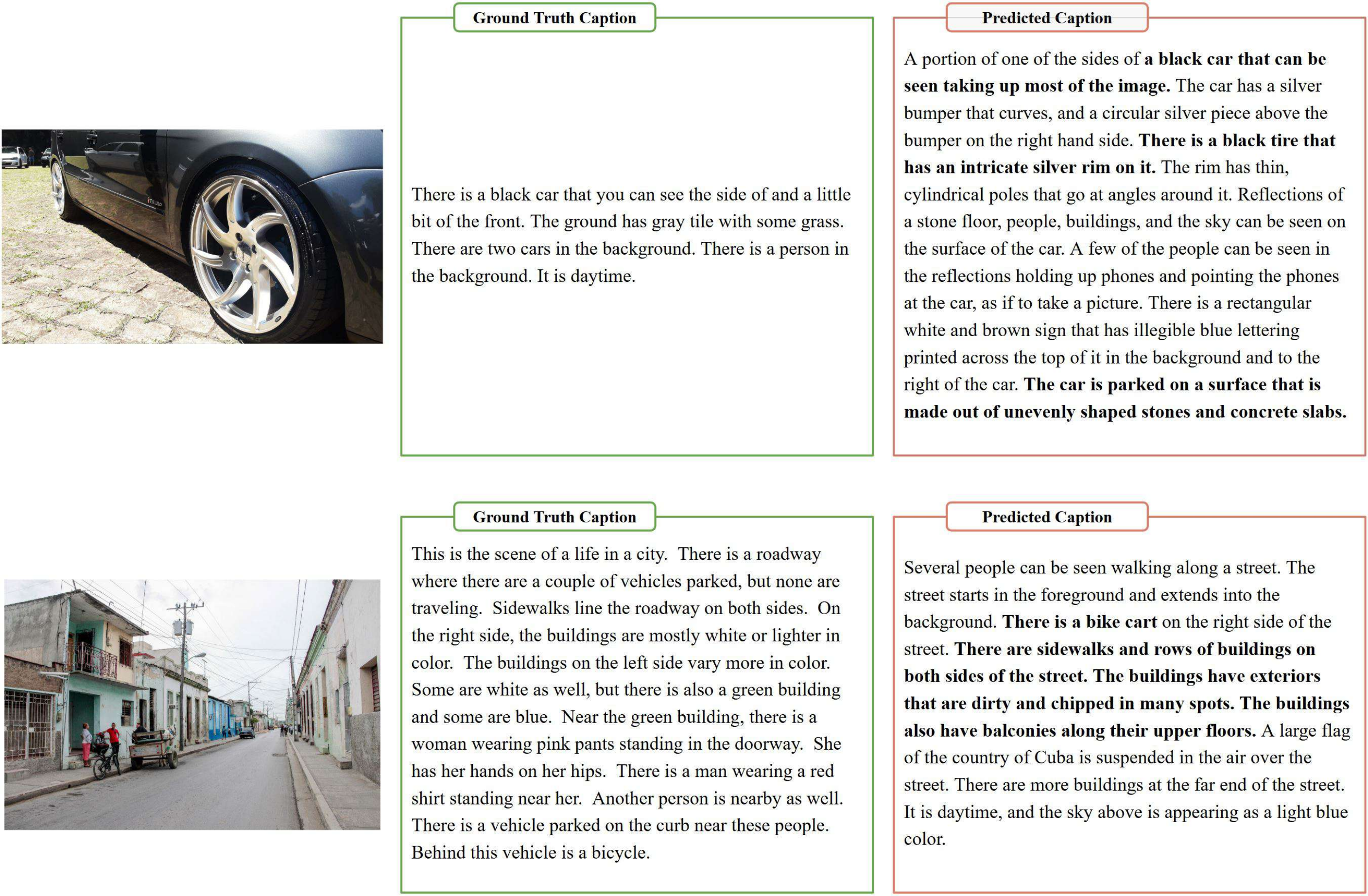}
\caption{Representative error cases from DCI~\cite{DCI} image-to-text retrieval. The predicted captions describe the images with high semantic similarity to ground truth, demonstrating the challenge of discriminating among multiple detailed descriptions that share substantial semantic overlap.}
\label{fig:error_analysis}
\end{figure*}

Similarly, for the second query about ``yellow flowers with tinges of pink" and ``yellow building is long", we observe that both baseline methods struggle with capturing the complete scene description. CLIP retrieves an image with yellow flowers and a building but does not precisely match the detailed description provided in the text, while Long-CLIP shows a building but lacks the distinctive yellow flowers described in the query.

In contrast, FAST-GOAL consistently retrieves more accurate images that closely match the comprehensive descriptions provided in the queries. For the first query, FAST-GOAL successfully identifies the correct image showing people walking past the grey brick building with the distinctive cafe signage. For the second query, FAST-GOAL retrieves an image that contains both the yellow building and the field of yellow flowers with pink tinges, demonstrating its ability to understand and match multiple detailed elements simultaneously. This superior performance stems from our method's enhanced ability to learn fine-grained correspondences between local image regions and specific textual phrases through the FLISM pipeline and TSL mechanism.

These qualitative findings complement our quantitative results, providing visual evidence that the improvements of FAST-GOAL in recall metrics translate to substantially enhanced retrieval quality in practical scenarios. The results demonstrate that while traditional methods achieve reasonable semantic alignment at a coarse level, the global-local learning approach of FAST-GOAL enables more precise understanding of fine-grained visual-textual relationships. This capability makes our method particularly effective for applications requiring accurate retrieval based on comprehensive and detailed descriptions.

\noindent {\bf{Error Analysis.}} To provide deeper insights into the challenges of lengthy text retrieval, we analyze representative error cases from our experiments. Figure~\ref{fig:error_analysis} presents retrieval errors from the DCI~\cite{DCI} dataset, illustrating the inherent difficulty of fine-grained discrimination in long caption retrieval tasks. The examples reveal the primary challenge: the text gallery contains multiple highly similar descriptions that differ only in subtle details.

In the first example, both the ground truth and predicted captions describe a black car with detailed observations about the surroundings. However, subtle differences in specific details such as tire rim descriptions and parking surface materials lead to retrieval errors. Similarly, in the second example, both captions describe a city street scene with buildings and people, yet differ in specific architectural details including building colors and balconies, as well as scene elements such as the bike cart. These errors occur because DCI captions provide detailed descriptions where multiple texts can semantically match the same image at a high level. The gallery contains descriptions that share substantial semantic overlap, creating challenging retrieval scenarios that require capturing every fine-grained detail for correct discrimination.

This analysis reveals important characteristics of lengthy text retrieval tasks. Unlike traditional short caption retrieval where semantic differences are often clear and substantial, lengthy descriptions introduce scenarios where multiple captions can accurately describe the same visual content with only minor variations in specific details. Successfully discriminating among such highly similar descriptions requires enhanced capability in understanding subtle semantic differences, presenting opportunities for future research directions in fine-grained multi-modal understanding.

\section{Conclusion}
In this paper, we have presented FAST-GOAL, a fast and efficient fine-tuning approach for enhancing the lengthy text understanding capability of CLIP. Our method has introduced two key components: Fast Local Image-Sentence Matching (FLISM) for efficient local correspondence extraction and Token Similarity-based Learning (TSL) for propagating attention from local elements to global representations. Furthermore, we have constructed GLIT100k, a dataset that provides both global image-lengthy caption pairs and context-derived fine-grained local pairs for multi-level supervision, enabling effective training without million-scale data requirements. By addressing the computational bottlenecks that have limited previous approaches, FAST-GOAL enables practical adaptation of vision-language models to detailed textual descriptions at scale efficiently.

\section{Acknowledgment}
This work was supported by the National Research Foundation of Korea(NRF) grant funded by the Korea government(MSIT)(RS-2024-00355008), the MSIT(Ministry of Science and ICT), Korea, under the Graduate School of Virtual Convergence support program(IITP-2024-RS-2024-00418847) supervised by the IITP(Institute for Information \& Communications Technology Planning \& Evaluation), and Culture, Sports and Tourism R\&D Program through the Korea Creative Content Agency grant funded by Ministry of Culture, Sports and Tourism in 2024 (Project Name : Developing Professionals for R\&D in Contents Production Based on Generative Ai and Cloud, Project Number : RS-2024-00352578, Contribution Rate: 30\%).

\bibliographystyle{IEEEtran} 
\bibliography{main}

@inproceedings{ALIGN,
  author={Jia, Chao and Yang, Yinfei and Xia, Ye and Chen, Yi-Ting and Parekh, Zarana and Pham, Hieu and Le, Quoc and Sung, Yun-Hsuan and Li, Zhen and Duerig, Tom},
  title={Scaling Up Visual and Vision-Language Representation Learning with Noisy Text Supervision},
  booktitle={Proceedings of the International Conference on Machine Learning},
  year={2021},
  pages={4904--4916}
}

@inproceedings{Florence,
  author={Yuan, Lu and Chen, Dongdong and Chen, Yi-Ling and Codella, Noel and Dai, Xiyang and Gao, Jianfeng and Hu, Houdong and Huang, Xuedong and Li, Boxin and Li, Chunyuan and others},
  title={Florence: A New Foundation Model for Computer Vision},
  booktitle={arXiv preprint arXiv:2111.11432},
  year={2021}
}

@inproceedings{CLIP,
  author={Radford, Alec and Kim, Jong Wook and Hallacy, Chris and Ramesh, Aditya and Goh, Gabriel and Agarwal, Sandhini and Sastry, Girish and Askell, Amanda and Mishkin, Pamela and Clark, Jack and others},
  title={Learning Transferable Visual Representations from Natural Language Supervision},
  booktitle={Proceedings of the International Conference on Machine Learning},
  year={2021},
  pages={8748--8763}
}

@inproceedings{COCO,
  author={Lin, Tsung-Yi and Maire, Michael and Belongie, Serge and Hays, James and Perona, Pietro and Ramanan, Deva and Doll{\'a}r, Piotr and Zitnick, C Lawrence},
  title={Microsoft {COCO}: Common Objects in Context},
  booktitle={Proceedings of the European Conference on Computer Vision},
  year={2014},
  pages={740--755}
}

@inproceedings{Flickr30k,
  author={Young, Peter and Lai, Alice and Hodosh, Micah and Hockenmaier, Julia},
  title={From Image Descriptions to Visual Denotations: New Similarity Metrics for Semantic Inference Over Event Descriptions},
  booktitle={Transactions of the Association for Computational Linguistics},
  year={2014},
  pages={67--78}
}

@inproceedings{DOCCI,
  author={Onoe, Yasumasa and Rane, Sunayana and Berger, Zachary and Bitton, Yonatan and Cho, Jaemin and Garg, Roopal and Ku, Alexander and Parekh, Zarana and Pont-Tuset, Jordi and Tanzer, Garrett and others},
  title={{DOCCI}: Descriptions of Connected and Contrasting Images},
  booktitle={Proceedings of the European Conference on Computer Vision},
  year={2024},
  pages={291--309}
}

@inproceedings{DCI,
  author={Urbanek, Jack and Bordes, Florian and Astolfi, Pietro and Williamson, Mary and Sharma, Vasu and Romero-Soriano, Adriana},
  title={A Picture is Worth More Than 77 Text Tokens: Evaluating {CLIP}-Style Models on Dense Captions},
  booktitle={Proceedings of the IEEE/CVF Conference on Computer Vision and Pattern Recognition},
  year={2024},
  pages={26700--26709}
}

@inproceedings{LongCLIP,
  author={Zhang, Beichen and Zhang, Pan and Dong, Xiaoyi and Zang, Yuhang and Wang, Jiaqi},
  title={Long-{CLIP}: Unlocking the Long-Text Capability of {CLIP}},
  booktitle={Proceedings of the European Conference on Computer Vision},
  year={2025},
  pages={310--325}
}

@inproceedings{SAM,
  author={Kirillov, Alexander and Mintun, Eric and Ravi, Nikhila and Mao, Hanzi and Rolland, Chloe and Gustafson, Laura and Xiao, Tete and Whitehead, Spencer and Berg, Alexander C and Lo, Wan-Yen and others},
  title={Segment Anything},
  booktitle={Proceedings of the IEEE/CVF International Conference on Computer Vision},
  year={2023},
  pages={4015--4026}
}

@inproceedings{ViTAA,
  author={Wang, Zhe and Fang, Zhiyuan and Wang, Jun and Yang, Yezhou},
  title={{ViTAA}: Visual-Textual Attributes Alignment in Person Search by Natural Language},
  booktitle={Proceedings of the European Conference on Computer Vision},
  year={2020},
  pages={402--420}
}

@inproceedings{CLOC,
  author={Chen, Hong-You and Lai, Zhengfeng and Zhang, Haotian and Wang, Xinze and Eichner, Marcin and You, Keen and Cao, Meng and Zhang, Bowen and Yang, Yinfei and Gan, Zhe},
  title={Contrastive Localized Language-Image Pre-Training},
  booktitle={Proceedings of the International Conference on Machine Learning},
  year={2025}
}

@inproceedings{OWLv2,
  author={Minderer, Matthias and Gritsenko, Alexey and Houlsby, Neil},
  title={Scaling Open-Vocabulary Object Detection},
  booktitle={Advances in Neural Information Processing Systems},
  year={2023},
  pages={72983--73007}
}

@inproceedings{GLIPv2,
  author={Zhang, Haotian and Zhang, Pengchuan and Hu, Xiaowei and Chen, Yen-Chun and Li, Liunian and Dai, Xiyang and Wang, Lijuan and Yuan, Lu and Hwang, Jenq-Neng and Gao, Jianfeng},
  title={{GLIPv2}: Unifying Localization and Vision-Language Understanding},
  booktitle={Advances in Neural Information Processing Systems},
  year={2022},
  pages={36067--36080}
}

@inproceedings{reid1,
  author={Tan, Wentan and Ding, Changxing and Jiang, Jiayu and Wang, Fei and Zhan, Yibing and Tao, Dapeng},
  title={Harnessing the Power of {MLLMs} for Transferable Text-to-Image Person {ReID}},
  booktitle={Proceedings of the IEEE/CVF Conference on Computer Vision and Pattern Recognition},
  year={2024},
  pages={17127--17137}
}

@inproceedings{reid2,
  author={Cui, Zhenyu and Zhou, Jiahuan and Wang, Xun and Zhu, Manyu and Peng, Yuxin},
  title={Learning Continual Compatible Representation for Re-indexing Free Lifelong Person Re-identification},
  booktitle={Proceedings of the IEEE/CVF Conference on Computer Vision and Pattern Recognition},
  year={2024},
  pages={16614--16623}
}

@inproceedings{reid3,
  author={Zheng, Liang and Zhang, Hengheng and Sun, Shaoyan and Chandraker, Manmohan and Yang, Yi and Tian, Qi},
  title={Person Re-identification in the Wild},
  booktitle={Proceedings of the IEEE/CVF Conference on Computer Vision and Pattern Recognition},
  year={2017},
  pages={1367--1376}
}

@inproceedings{reid4,
  author={Zhong, Zhun and Zheng, Liang and Cao, Donglin and Li, Shaozi},
  title={Re-ranking Person Re-identification with k-Reciprocal Encoding},
  booktitle={Proceedings of the IEEE/CVF Conference on Computer Vision and Pattern Recognition},
  year={2017},
  pages={1318--1327}
}

@inproceedings{ViT,
  author={Dosovitskiy, Alexey and Beyer, Lucas and Kolesnikov, Alexander and Weissenborn, Dirk and Zhai, Xiaohua and Unterthiner, Thomas and Dehghani, Mostafa and Minderer, Matthias and Heigold, Georg and Gelly, Sylvain and others},
  title={An Image is Worth 16x16 Words: Transformers for Image Recognition at Scale},
  booktitle={Proceedings of the International Conference on Learning Representations},
  year={2021}
}

@inproceedings{sharegpt4v,
  author={Chen, Lin and Li, Jinsong and Dong, Xiaoyi and Zhang, Pan and He, Conghui and Wang, Jiaqi and Zhao, Feng and Lin, Dahua},
  title={{ShareGPT4V}: Improving Large Multi-Modal Models with Better Captions},
  booktitle={Proceedings of the European Conference on Computer Vision},
  year={2024},
  pages={370--387}
}

@inproceedings{GOAL,
  author={Choi, Hyungyu and Jang, Young Kyun and Eom, Chanho},
  title={{GOAL}: Global-Local Object Alignment Learning},
  booktitle={Proceedings of the IEEE/CVF Conference on Computer Vision and Pattern Recognition},
  year={2025},
  pages={4070--4079}
}

@inproceedings{FG-CLIP,
  author={Xie, Chunyu and Wang, Bin and Kong, Fanjing and Li, Jincheng and Liang, Dawei and Zhang, Gengshen and Leng, Dawei and Yin, Yuhui},
  title={{FG-CLIP}: Fine-Grained Visual and Textual Alignment},
  booktitle={Proceedings of the International Conference on Machine Learning},
  year={2025}
}

@misc{LLaVA-NEXT,
  author={Liu, Haotian and Li, Chunyuan and Li, Yuheng and Li, Bo and Zhang, Yuanhan and Shen, Sheng and Lee, Yong Jae},
  title={{LLaVA-NeXT}: Improved Reasoning, {OCR}, and World Knowledge},
  year={2024}
}

@inproceedings{YOLOS,
  author={Fang, Yuxin and Liao, Bencheng and Wang, Xinggang and Fang, Jiemin and Qi, Jiyang and Wu, Rui and Niu, Jianwei and Liu, Wenyu},
  title={You Only Look at One Sequence: Rethinking Transformer in Vision through Object Detection},
  booktitle={Advances in Neural Information Processing Systems},
  year={2021},
  pages={26183--26197}
}

@inproceedings{EVA-CLIP,
  author={Sun, Quan and Fang, Yuxin and Wu, Ledell and Wang, Xinlong and Cao, Yue},
  title={{EVA-CLIP}: Improved Training Techniques for {CLIP} at Scale},
  booktitle={arXiv preprint arXiv:2303.15389},
  year={2023}
}

@inproceedings{FineCLIP,
  author={Jing, Dong and He, Xiaolong and Luo, Yutian and Fei, Nanyi and Wei, Wei and Zhao, Huiwen and Lu, Zhiwu and others},
  title={{FineCLIP}: Self-Distilled Region-Based {CLIP} for Better Fine-Grained Understanding},
  booktitle={Advances in Neural Information Processing Systems},
  year={2024},
  pages={27896--27918}
}

@inproceedings{ALBEF,
  author={Li, Junnan and Selvaraju, Ramprasaath and Gotmare, Akhilesh and Joty, Shafiq and Xiong, Caiming and Hoi, Steven Chu Hong},
  title={Align Before Fuse: Vision and Language Representation Learning with Momentum Distillation},
  booktitle={Advances in Neural Information Processing Systems},
  year={2021},
  pages={9694--9705}
}

@inproceedings{LiT,
  author={Zhai, Xiaohua and Wang, Xiao and Mustafa, Basil and Steiner, Andreas and Keysers, Daniel and Kolesnikov, Alexander and Beyer, Lucas},
  title={{LiT}: Zero-Shot Transfer with Locked-Image Text Tuning},
  booktitle={Proceedings of the IEEE/CVF Conference on Computer Vision and Pattern Recognition},
  year={2022},
  pages={18123--18133}
}

@inproceedings{Llip,
  author={Lavoie, Samuel and Kirichenko, Polina and Ibrahim, Mark and Assran, Mahmoud and Wilson, Andrew Gordon and Courville, Aaron and Ballas, Nicolas},
  title={Modeling Caption Diversity in Contrastive Vision-Language Pretraining},
  booktitle={Proceedings of the International Conference on Machine Learning},
  year={2024},
  pages={26070--26084}
}

@inproceedings{SigLIP,
  author={Zhai, Xiaohua and Mustafa, Basil and Kolesnikov, Alexander and Beyer, Lucas},
  title={Sigmoid Loss for Language Image Pre-Training},
  booktitle={Proceedings of the IEEE/CVF International Conference on Computer Vision},
  year={2023},
  pages={11975--11986}
}

@inproceedings{AlexNet,
  author={Krizhevsky, Alex and Sutskever, Ilya and Hinton, Geoffrey E},
  title={{ImageNet} Classification with Deep Convolutional Neural Networks},
  booktitle={Advances in Neural Information Processing Systems},
  year={2012},
  pages={1097--1105}
}

@inproceedings{VGGNet,
  author={Simonyan, Karen and Zisserman, Andrew},
  title={Very Deep Convolutional Networks for Large-Scale Image Recognition},
  booktitle={Proceedings of the International Conference on Learning Representations},
  year={2014}
}

@inproceedings{ResNet,
  author={He, Kaiming and Zhang, Xiangyu and Ren, Shaoqing and Sun, Jian},
  title={Deep Residual Learning for Image Recognition},
  booktitle={Proceedings of the IEEE/CVF Conference on Computer Vision and Pattern Recognition},
  year={2016},
  pages={770--778}
}

@inproceedings{Inception,
  author={Szegedy, Christian and Vanhoucke, Vincent and Ioffe, Sergey and Shlens, Jon and Wojna, Zbigniew},
  title={Rethinking the Inception Architecture for Computer Vision},
  booktitle={Proceedings of the IEEE/CVF Conference on Computer Vision and Pattern Recognition},
  year={2016},
  pages={2818--2826}
}

@inproceedings{DeViSE,
  author={Frome, Andrea and Corrado, Greg S and Shlens, Jon and Bengio, Samy and Dean, Jeff and Ranzato, Marc'Aurelio and Mikolov, Tomas},
  title={{DeViSE}: A Deep Visual-Semantic Embedding Model},
  booktitle={Advances in Neural Information Processing Systems},
  year={2013},
  pages= {2121--2129}
}

@inproceedings{DeViSE+,
  author={Karpathy, Andrej and Fei-Fei, Li},
  title={Deep Visual-Semantic Alignments for Generating Image Descriptions},
  booktitle={Proceedings of the IEEE/CVF Conference on Computer Vision and Pattern Recognition},
  year={2015},
  pages={3128--3137}
}

@inproceedings{LLaVA,
  author={Liu, Haotian and Li, Chunyuan and Wu, Qingyang and Lee, Yong Jae},
  title={Visual Instruction Tuning},
  booktitle={Advances in Neural Information Processing Systems},
  year={2023},
  pages={34892--34916}
}

@inproceedings{Qwen,
  author={Bai, Jinze and Bai, Shuai and Chu, Yunfei and Cui, Zeyu and Dang, Kai and Deng, Xiaodong and Fan, Yang and Ge, Wenbin and Han, Yu and Huang, Fei and others},
  title={Qwen Technical Report},
  booktitle={arXiv preprint arXiv:2309.16609},
  year={2023}
}

@inproceedings{Gemini,
  author={Team, Gemini and Anil, Rohan and Borgeaud, Sebastian and Alayrac, Jean-Baptiste and Yu, Jiahui and Soricut, Radu and Schalkwyk, Johan and Dai, Andrew M and Hauth, Anja and Millican, Katie and others},
  title={Gemini: A Family of Highly Capable Multimodal Models},
  booktitle={arXiv preprint arXiv:2312.11805},
  year={2023}
}

@inproceedings{InternVL,
  author={Chen, Zhe and Wu, Jiannan and Wang, Wenhai and Su, Weijie and Chen, Guo and Xing, Sen and Zhong, Muyan and Zhang, Qinglong and Zhu, Xizhou and Lu, Lewei and others},
  title={{InternVL}: Scaling Up Vision Foundation Models and Aligning for Generic Visual-Linguistic Tasks},
  booktitle={Proceedings of the IEEE/CVF Conference on Computer Vision and Pattern Recognition},
  year={2024},
  pages={24185--24198}
}

@inproceedings{PCA,
  author={Hotelling, Harold},
  title={Analysis of a Complex of Statistical Variables into Principal Components},
  booktitle={Journal of Educational Psychology},
  year={1933},
  pages={417--441}
}

@inproceedings{FineLIP,
  author={Asokan, Mothilal and Wu, Kebin and Albreiki, Fatima},
  title={{FineLIP}: Extending {CLIP}'s Reach via Fine-Grained Alignment with Longer Text Inputs},
  booktitle={Proceedings of the IEEE/CVF Conference on Computer Vision and Pattern Recognition},
  year={2025},
  pages={14495--14504}
}

@InProceedings{DenseVLM,
    author    = {Li, Yunheng and Li, Yuxuan and Zeng, Quan-Sheng and Wang, Wenhai and Hou, Qibin and Cheng, Ming-Ming},
    title     = {Unbiased Region-Language Alignment for Open-Vocabulary Dense Prediction},
    booktitle = {Proceedings of the IEEE/CVF International Conference on Computer Vision (ICCV)},
    year      = {2025},
    pages     = {23795-23805}
}

@article{retrieval1,
  title={Intelligent Video Surveillance System with Abnormal Behavior Recognition and Metadata Retrieval},
  author={Kim, Hyungtae and Shin, Joongchol and Park, Seokmok and Paik, Joonki},
  year={2023},
  publisher={Preprints}
}

@article{retrieval2,
  title={Tourism Image Retrieval Method Based on Deep Residual Shrinkage Network},
  author={Zhao, Renbi},
  journal={IEIE Transactions on Smart Processing \& Computing},
  volume={14},
  number={3},
  pages={419--429},
  year={2025}
}

\vspace{10cm}

\begin{IEEEbiography}[{{\includegraphics[width=1in,height=1.3in,clip,keepaspectratio]{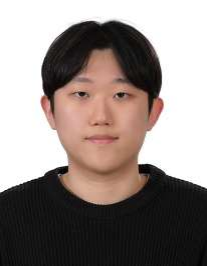}}}]{Hyungyu Choi}
is a researcher at the Graduate School of Advanced Imaging Science, Multimedia, and Film (GSAIM) at Chung-Ang University in Seoul, Korea. He received his B.S. degree in Mathematics from Chung-Ang University in 2025. He is currently conducting research as a member of the Perceptual AI LAB under the supervision of Professor Chanho Eom. His research interests focus on vision-language models, particularly in multi-modal understanding and cross-modal retrieval.
\end{IEEEbiography}

\vspace{-10cm}

\begin{IEEEbiography}[{{\includegraphics[width=1in,height=1.3in,clip,keepaspectratio]{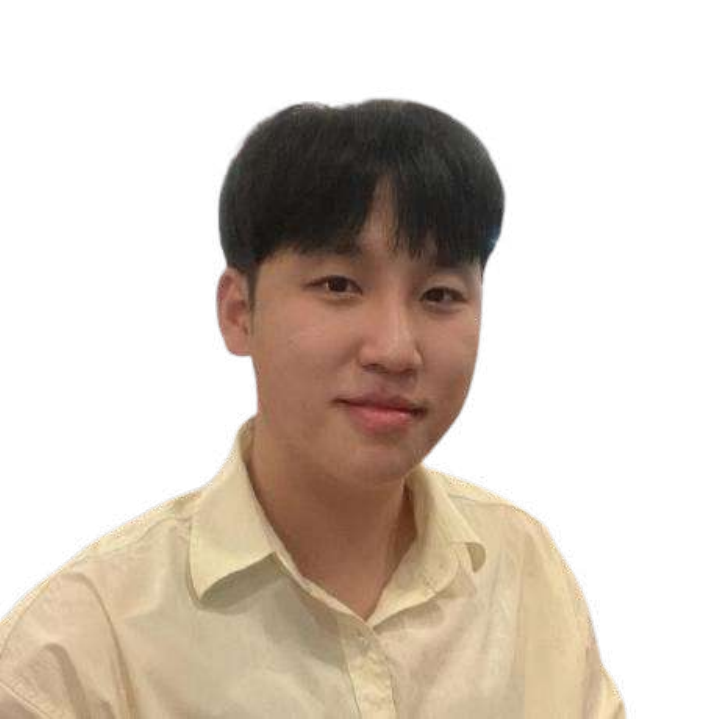}}}]{Young Kyun Jang}is a Research Scientist who received his Ph.D. in Electrical and Computer Engineering from Intelligent Signal Processing Lab of Seoul National University. His research interests include multi-modal representation learning, efficient retrieval, and foundational (large language) models.
\end{IEEEbiography}

\vspace{-10cm}

\begin{IEEEbiography}[{{\includegraphics[width=1in,height=1.3in,clip,keepaspectratio]{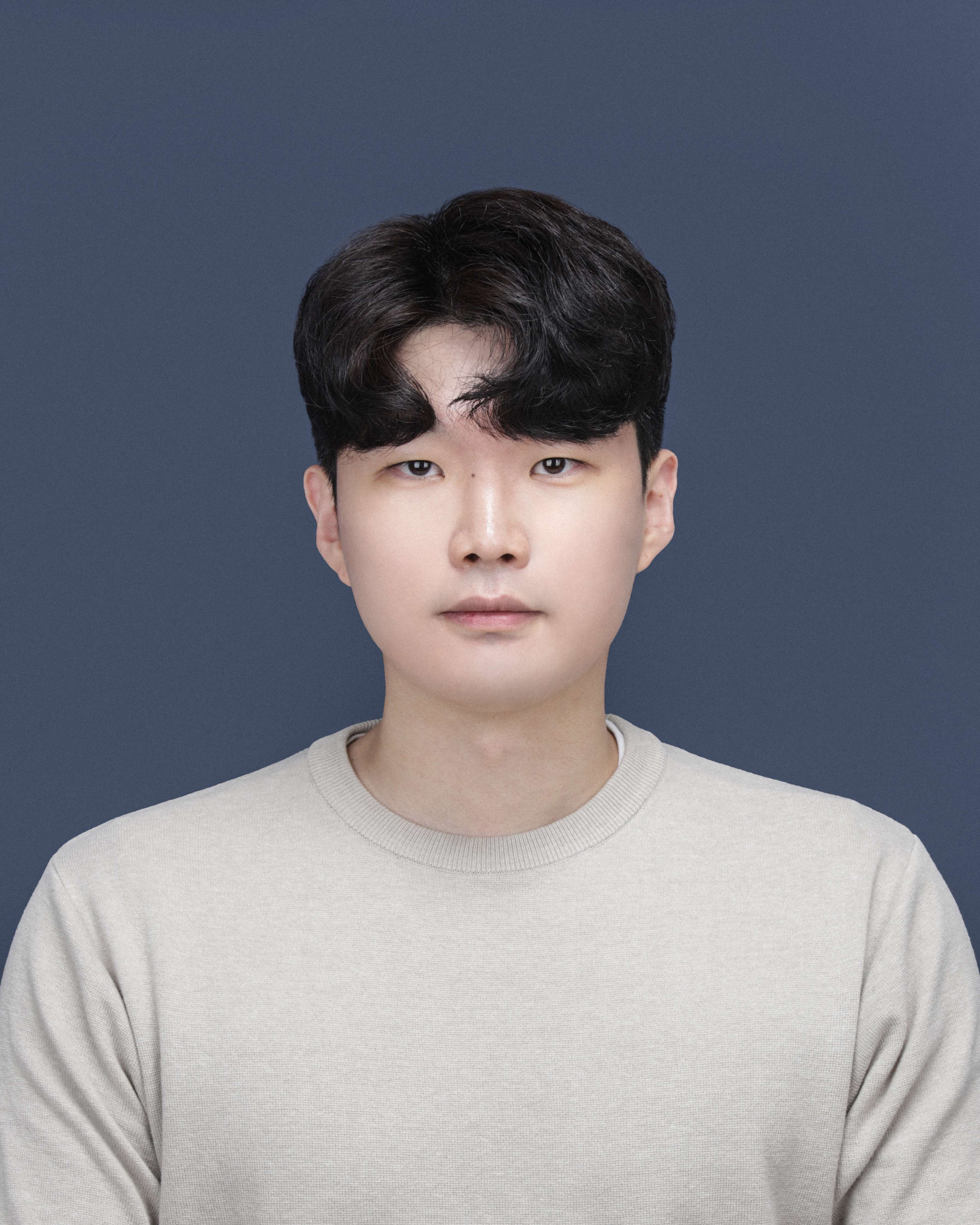}}}]{Chanho Eom}
is an Assistant Professor at the Graduate School of Advanced Imaging Science, Multimedia, and Film (GSAIM) at Chung-Ang University in Seoul, Korea. He received his B.S. and Ph.D. degrees in Electrical and Electronic Engineering from Yonsei University in 2017 and 2023, respectively. He previously worked as a researcher at the Samsung Advanced Institute of Technology (SAIT). His research interests include computer vision and deep learning, particularly in retrieval, person re-identification, and video analysis, both in theory~and~applications.
\end{IEEEbiography}



\section*{Supplementary material}

# Supplementary Material

TABLE IX
COMPLETE END-TO-END TIME COMPARISON ON DOCCI DATASET. FAST-GOAL ACHIEVES 12.7× FASTER PREPROCESSING AND 6.2-8.2× FASTER TOTAL TIME WHILE MAINTAINING IDENTICAL TRAINING EFFICIENCY.

| Method | Preprocessing | Training | Total Time |
|---|---|---|---|
| GOAL | 1,239.7 min | 60-120 min | 1,299.7-1,359.7 min |
| FAST-GOAL | 97.63 min | 60-120 min | 157.63-217.63 min |
| Speedup | 12.7× | 1× | 6.2-8.2× |

TABLE X
COMPLETE PARAMETER COMPARISON. FAST-GOAL ACHIEVES 2.2× PARAMETER REDUCTION COMPARED TO GOAL WHILE MAINTAINING REGION EXTRACTION QUALITY.

| Method | Image Encoder | Text Encoder | Region Extraction | Total Parameters |
|---|---|---|---|---|
| GOAL | 149.6M (ViT-B/16) | 149.6M (ViT-B/16) | 641.1M (SAM) | 940.3M |
| FAST-GOAL | 149.6M (ViT-B/16) | 149.6M (ViT-B/16) | 127.8M (YOLOS) | 426.9M |
| Reduction | 1× | 1× | 5× | 2.2× |

## A. Computational Efficiency Analysis

**End-to-End Time Efficiency.** Table IX presents the complete end-to-end time comparison between GOAL [1] and FAST-GOAL, demonstrating where efficiency gains occur in our pipeline. The ”Fast” in FAST-GOAL refers specifically to preprocessing stage efficiency. FLISM preprocessing is performed before fine-tuning to generate local pairs from the dataset, while the actual training process follows the same architecture as GOAL. The results show that while both methods require identical training time of 60-120 minutes, FAST-GOAL achieves dramatic preprocessing time reduction from 1,239.7 minutes to 97.63 minutes on DOCCI [2], representing a 12.7× speedup. This translates to total time reduction from 1,299.7-1,359.7 minutes to 157.63-217.63 minutes, achieving a 6.2-8.2× overall speedup.

This efficiency advantage directly addresses the scalability bottleneck in long text understanding. The SAM [3]-based preprocessing in GOAL makes large-scale dataset construction prohibitively expensive, requiring approximately 1,378 hours (57 days) to process 100k images. In contrast, FAST-GOAL's YOLO [4]-based approach enables construction of GLIT100k with 100,647 global-local pairs in practical timeframes, removing the computational bottleneck that constrained previous methods. Importantly, this faster preprocessing does not compromise quality, as demonstrated in our ablation studies where YOLO-based extraction achieves comparable performance to SAM-based approaches.

**Parameter Efficiency and Architectural Design.** Table X presents the complete parameter comparison between GOAL and FAST-GOAL. Our choice of YOLOS over SAM is strategically motivated by parameter efficiency. GOAL requires 940.3M total parameters, combining CLIP's [5] image encoder (149.6M), text encoder (149.6M), and SAM for region extraction (641.1M). In contrast, FAST-GOAL requires only 426.9M parameters by replacing SAM with YOLOS (127.8M), achieving a 2.2× parameter reduction. This significant reduction, combined with object detection's focused nature compared to dense segmentation, enables the 12.7× preprocessing speedup while maintaining semantic region identification quality.

## B. Additional Experimental Analysis

**Experiments with ViT-L/14 Backbone.** To demonstrate the scalability of our approach across different model sizes, we conduct comprehensive experiments using ViT-L/14 [6] backbone following the same evaluation protocol as our ViT-B/16 experiments. Tables XI and XII present the results on long caption and short caption domains, respectively. Table XI shows the performance on DOCCI [2] and DCI [7] datasets. FAST-GOAL consistently achieves superior performance across both datasets. On DOCCI, FAST-GOAL attains 78.55% text-to-image R@1 and 78.61% image-to-text R@1, demonstrating substantial improvements over available baselines. On DCI, FAST-GOAL achieves 59.73% text-to-image R@1 and 60.76% image-to-text R@1, outperforming Long-CLIP [8] with 54.31% and 46.53%, and GOAL [1] with 54.66% and 54.67%. These results validate that our method effectively scales to larger model architectures while maintaining its advantages in long-text understanding. Table XII demonstrates performance on MSCOCO [9] and Flickr30k [10] datasets using standard R@1/5/10/15 metrics. Despite being optimized for long-text scenarios, FAST-GOAL maintains competitive performance on short caption domains. On MSCOCO, FAST-GOAL achieves 48.53% text-to-image R@1 and 66.58% image-to-text R@1, comparable to EVA-CLIP [11] with 48.14% and 63.96%. On Flickr30k, FAST-GOAL attains 76.72% text-to-image R@1 and 92.50% image-to-text R@1. These results confirm that our approach demonstrates strong generalizability across different caption lengths and model scales, validating the robustness of our global-local alignment strategy.

**Comparison with recent CLIP-based models.** Table XIII presents performance comparisons with recent CLIP-based models that employ different data scales. This comparison highlights the data efficiency advantage of our approach, particularly on long caption understanding tasks. SigLIPv2 [12] and MobileCLIPv2 [13] are pretrained on massive-scale datasets of 10 billion and 12 billion samples respectively, representing data scales that are 100-120× larger than CLIP's pretraining data. While these models demonstrate strong performance across various benchmarks, their reliance on such massive data requirements creates significant barriers for practical adoption and research accessibility. In contrast, FAST-GOAL builds upon standard pretrained CLIP trained on 400M samples and requires only an additional 100k samples from

TABLE XI
PERFORMANCE COMPARISON ON LONG CAPTION DOMAIN DATASETS (DOCCI AND DCI) USING VIT-L/14 BACKBONE. RESULTS SHOW RECALL@K SCORES FOR TEXT TO IMAGE AND IMAGE TO TEXT RETRIEVAL TASKS. THE BEST PERFORMANCE IS SHOWN IN **BOLD**, AND THE SECOND-BEST PERFORMANCE IS UNDERLINED.

| Dataset | Method | Backbone | Text to Image Recall@K | | | | Image to Text Recall@K | | | |
|---|---|---|---|---|---|---|---|---|---|---|
| | | | R@1 | R@5 | R@25 | R@50 | R@1 | R@5 | R@25 | R@50 |
| DOCCI | CLIP | ViT-L/14 | 60.51 | 86.31 | 96.67 | 98.45 | 57.75 | 83.53 | 95.33 | 97.73 |
| | EVA-CLIP | | 73.08 | 93.10 | 98.25 | 99.27 | 72.12 | 92.16 | 98.47 | 99.35 |
| | Long-CLIP | | 78.84 | 95.25 | 99.19 | 99.59 | 66.82 | 91.90 | 99.04 | 99.82 |
| | FineCLIP | | 66.29 | 88.37 | 97.24 | 98.75 | 56.24 | 82.75 | 95.43 | 97.90 |
| | GOAL | | **79.04** | 95.78 | **99.55** | **99.84** | **79.16** | 95.96 | 99.61 | **99.90** |
| | FAST-GOAL | | 78.55 | **95.86** | 99.51 | 99.80 | 78.61 | **96.24** | **99.67** | **99.90** |
| DCI | CLIP | ViT-L/14 | 31.18 | 50.34 | 68.84 | 76.15 | 33.37 | 51.51 | 68.25 | 75.70 |
| | EVA-CLIP | | 45.88 | 64.96 | 79.56 | 84.87 | 44.70 | 64.75 | 79.94 | 85.41 |
| | Long-CLIP | | 54.31 | 72.53 | 85.04 | 89.06 | 46.53 | 72.64 | 86.48 | 90.38 |
| | FineCLIP | | 42.12 | 62.01 | 78.11 | 83.94 | 35.54 | 57.04 | 75.09 | 81.49 |
| | GOAL | | 54.66 | 74.57 | 86.29 | 90.54 | 54.67 | 74.86 | 87.12 | 90.94 |
| | FAST-GOAL | | **59.73** | **77.23** | **88.21** | **91.45** | **60.76** | **78.93** | **90.42** | **93.61** |

TABLE XII
PERFORMANCE COMPARISON ON SHORT CAPTION DOMAIN DATASETS (MSCOCO AND FLICKR30K) USING VIT-L/14 BACKBONE WITH STANDARD EVALUATION METRICS. RESULTS SHOW RECALL@K SCORES FOR TEXT TO IMAGE AND IMAGE TO TEXT RETRIEVAL TASKS. THE BEST PERFORMANCE IS SHOWN IN **BOLD**, AND THE SECOND-BEST PERFORMANCE IS UNDERLINED. NOTE THAT FINECLIP ON MSCOCO IS FINE-TUNED ON MSCOCO ITSELF, THUS IT IS NOT A FAIR ZERO-SHOT COMPARISON.

| Dataset | Method | Backbone | Text to Image Recall@K | | | | Image to Text Recall@K | | | |
|---|---|---|---|---|---|---|---|---|---|---|
| | | | R@1 | R@5 | R@10 | R@15 | R@1 | R@5 | R@10 | R@15 |
| MSCOCO | CLIP | ViT-L/14 | 37.29 | 61.82 | 71.64 | 77.46 | 57.68 | 80.20 | 87.52 | 90.84 |
| | EVA-CLIP | | 48.14 | 71.72 | 80.10 | 84.46 | 63.96 | 84.68 | 91.04 | 93.66 |
| | Long-CLIP | | 46.96 | 71.89 | 80.64 | 85.31 | 63.16 | 84.52 | 90.60 | 93.44 |
| | GOAL | | 46.29 | 70.85 | 79.67 | 84.40 | 66.50 | 86.04 | 91.74 | 94.36 |
| | FAST-GOAL | | **48.53** | **73.22** | **82.12** | **86.30** | **66.58** | **86.78** | **92.40** | **94.72** |
| Flickr30k | CLIP | ViT-L/14 | 65.38 | 87.36 | 92.24 | 94.68 | 86.40 | 97.50 | 99.30 | 99.40 |
| | EVA-CLIP | | **77.40** | **93.98** | **97.18** | **98.16** | 91.10 | 99.10 | **99.80** | 99.80 |
| | Long-CLIP | | 76.22 | 93.54 | 96.26 | 97.40 | 90.00 | 98.90 | 99.50 | **99.90** |
| | FineCLIP | | 72.68 | 91.56 | 95.58 | 96.78 | 80.80 | 95.90 | 97.60 | 98.70 |
| | GOAL | | 74.76 | 92.66 | 95.88 | 97.22 | 90.80 | 98.80 | 99.50 | 99.70 |
| | FAST-GOAL | | 76.72 | 93.72 | 96.74 | 97.60 | **92.50** | **99.40** | 99.70 | 99.80 |

GLIT100k for fine-tuning, making it substantially more accessible for researchers with limited computational resources.

Despite this dramatic difference in data scale, FAST-GOAL achieves competitive or superior performance on long caption understanding tasks. Most notably, on DCI, FAST-GOAL achieves 56.62% text-to-image R@1 and 56.74% image-to-text R@1, outperforming all compared methods including MobileCLIPv2 with 53.04% and 55.06% despite its 120× larger pretraining scale. On DOCCI, FAST-GOAL attains 74.27% text-to-image R@1 and 74.67% image-to-text R@1, demonstrating strong performance with only 100k fine-tuning samples compared to the billions of samples required by other approaches.

TinyCLIP [14], which shares CLIP's 400M pretraining scale, shows limited capability on lengthy descriptions with 34.77% on DCI, highlighting that pretraining scale alone is insufficient without targeted fine-tuning strategies. On short caption datasets where massive pretraining provides advantages, FAST-GOAL maintains reasonable performance despite being optimized for long-text scenarios. These results demonstrate that strategic fine-tuning with well-designed global-local alignment and carefully curated datasets can achieve strong long-text understanding without requiring billion-scale pretraining, offering a practical and efficient approach for extending vision-language models to detailed textual descriptions.

**Analysis of Hard Negative Mining.** Table XIV presents experimental exploration of hard negative mining strategies for our framework. We implement a two-stage training approach where after training FAST-GOAL with our standard pipeline, we perform additional fine-tuning using only in-image hard

TABLE XIII
PERFORMANCE COMPARISON WITH RECENT LIGHTWEIGHT AND EFFICIENT CLIP MODELS. RESULTS DEMONSTRATE THAT FAST-GOAL ACHIEVES COMPETITIVE OR SUPERIOR PERFORMANCE ON LONG CAPTION UNDERSTANDING THROUGH EFFICIENT FINE-TUNING. THE BEST PERFORMANCE IS SHOWN IN **BOLD**, AND THE SECOND-BEST PERFORMANCE IS UNDERLINED.

| Dataset | Method | Seen Data | Text to Image Recall@K | | | | Image to Text Recall@K | | | |
|---|---|---|---|---|---|---|---|---|---|---|
| | | | R@1 | R@5 | R@25 | R@50 | R@1 | R@5 | R@25 | R@50 |
| DOCCI | SigLIPv2 | 10B | 49.24 | 75.73 | 90.76 | 94.71 | 48.39 | 74.41 | 90.49 | 94.82 |
| | MobileCLIPv2 | 12B | **79.65** | **95.29** | 99.10 | 99.59 | **78.00** | **94.90** | 98.98 | 99.61 |
| | TinyCLIP | 400M | 59.10 | 83.90 | 95.43 | 97.94 | 58.88 | 83.63 | 95.35 | 97.82 |
| | FAST-GOAL | 400M+100k | 74.27 | 93.76 | **99.27** | **99.71** | 74.67 | 94.71 | **99.24** | **99.75** |
| DCI | SigLIPv2 | 10B | 29.62 | 49.80 | 67.82 | 75.34 | 29.64 | 48.61 | 67.73 | 75.11 |
| | MobileCLIPv2 | 12B | 53.04 | 72.20 | 84.81 | 89.09 | 55.06 | 73.61 | 86.16 | 90.00 |
| | TinyCLIP | 400M | 34.77 | 55.21 | 72.81 | 79.45 | 36.52 | 56.41 | 74.35 | 81.10 |
| | FAST-GOAL | 400M+100k | **56.62** | **75.26** | **87.75** | **91.08** | **56.74** | **75.99** | **88.71** | **92.46** |
| MSCOCO | SigLIPv2 | 10B | 49.28 | 73.36 | 90.33 | 95.08 | 65.90 | 86.62 | 96.76 | 98.78 |
| | MobileCLIPv2 | 12B | **49.63** | **74.92** | **91.58** | **96.26** | **67.44** | **86.76** | **97.00** | **99.10** |
| | TinyCLIP | 400M | 34.02 | 59.75 | 83.31 | 91.17 | 49.76 | 74.78 | 92.36 | 96.40 |
| | FAST-GOAL | 400M+100k | 42.81 | 68.61 | 88.54 | 94.64 | 61.00 | 82.84 | 95.34 | 98.32 |
| Flickr30k | SigLIPv2 | 10B | 74.24 | 91.48 | 97.76 | 98.94 | 90.10 | 98.30 | **100.00** | **100.00** |
| | MobileCLIPv2 | 12B | **76.60** | **93.22** | **98.54** | **99.20** | **90.40** | **98.40** | **100.00** | **100.00** |
| | TinyCLIP | 400M | 59.68 | 83.56 | 95.00 | 97.32 | 77.50 | 92.80 | 98.70 | 99.20 |
| | FAST-GOAL | 400M+100k | 72.50 | 91.34 | 97.92 | 98.80 | 86.80 | 97.50 | **100.00** | **100.00** |

TABLE XIV
ABLATION STUDY ON HARD NEGATIVE MINING STRATEGY USING VIT-B/16 BACKBONE. FAST-GOAL + HARD NEG REPRESENTS TWO-STAGE TRAINING WHERE THE SECOND STAGE USES ONLY IN-IMAGE LOCAL PAIRS AS HARD NEGATIVES. THE BEST PERFORMANCE IS SHOWN IN **BOLD**.

| Dataset | Method | Text to Image Recall@K | | | | Image to Text Recall@K | | | |
|---|---|---|---|---|---|---|---|---|---|
| | | R@1 | R@5 | R@25 | R@50 | R@1 | R@5 | R@25 | R@50 |
| DOCCI | FAST-GOAL + Hard Neg | 51.00 | 77.65 | 93.18 | 96.45 | 54.61 | 80.92 | 94.24 | 97.25 |
| | FAST-GOAL | **74.27** | **93.76** | **99.27** | **99.71** | **74.67** | **94.71** | **99.24** | **99.75** |
| DCI | FAST-GOAL + Hard Neg | 40.28 | 61.05 | 78.10 | 83.75 | 38.56 | 60.21 | 77.94 | 84.48 |
| | FAST-GOAL | **56.62** | **75.26** | **87.75** | **91.08** | **56.74** | **75.99** | **88.71** | **92.46** |
| MSCOCO | FAST-GOAL + Hard Neg | 34.30 | 59.46 | 82.44 | 90.50 | 48.70 | 73.68 | 91.36 | 95.56 |
| | FAST-GOAL | **42.81** | **68.61** | **88.54** | **94.64** | **61.00** | **82.84** | **95.34** | **98.32** |
| Flickr30k | FAST-GOAL + Hard Neg | 62.62 | 85.46 | 95.86 | 97.92 | 78.90 | 94.50 | 99.40 | 99.80 |
| | FAST-GOAL | **72.50** | **91.34** | **97.92** | **98.80** | **86.80** | **97.50** | **100.00** | **100.00** |

negatives. Specifically, we construct the contrastive learning matrix using only the multiple local pairs extracted from the same image, forcing the model to distinguish between different regions and their corresponding sentences within each sample.

The results show significant performance degradation across all datasets when applying hard negative mining. On DOCCI, text-to-image R@1 decreases from 74.27% to 51.00%. Similar patterns appear on DCI with performance dropping from 56.62% to 40.28%, on MSCOCO from 42.81% to 34.30%, and on Flickr30k from 72.50% to 62.62%. This substantial decline can be explained by the matching confidence distribution in our FLISM pipeline. The pipeline selects only the highest similarity local pair precisely because it represents the most confident and accurate correspondence. Other local pairs from the same image, despite being from the same source, have lower matching confidence and may represent less reliable region-sentence associations. Training with these lower-confidence pairs as hard negatives appears to introduce inconsistent supervision signals that affect model performance. These results suggest that effective hard negative mining in the context of local region-sentence matching requires careful consideration of matching confidence and correspondence quality.

**Comparison with Sentence-Level Averaging.** We examine an alternative approach for handling lengthy captions: splitting them into individual sentences, computing CLIP embeddings for each sentence separately, then averaging these embeddings for retrieval. Table XV presents results comparing this strategy against standard CLIP and FAST-GOAL. The results reveal fundamental limitations of simple averaging strategies. Sentence-level averaging achieves only 39.00% text-to-image R@1 on DOCCI and 19.48% on DCI, dramatically

TABLE XV
COMPARISON OF DIFFERENT APPROACHES FOR HANDLING LENGTHY TEXT USING VIT-B/16 BACKBONE. CLIP (SENTENCE-AVG) SPLITS CAPTIONS INTO SENTENCES AND AVERAGES THEIR EMBEDDINGS, WHILE CLIP (STANDARD) USES POSITIONAL EMBEDDING INTERPOLATION TO HANDLE 248 TOKENS. THE BEST PERFORMANCE IS SHOWN IN **BOLD**.

| Dataset | Method | Text to Image Recall@K | | | | Image to Text Recall@K | | | |
|---|---|---|---|---|---|---|---|---|---|
| | | R@1 | R@5 | R@25 | R@50 | R@1 | R@5 | R@25 | R@50 |
| DOCCI | CLIP (sentence-avg) | 39.00 | 68.82 | 89.20 | 94.18 | 40.88 | 71.37 | 91.14 | 95.84 |
| | CLIP (standard) | 58.53 | 85.57 | 97.16 | 98.82 | 58.80 | 85.18 | 96.73 | 98.45 |
| | FAST-GOAL | **74.27** | **93.76** | **99.27** | **99.71** | **74.67** | **94.71** | **99.24** | **99.75** |
| DCI | CLIP (sentence-avg) | 19.48 | 35.78 | 54.64 | 63.56 | 19.63 | 37.67 | 58.23 | 66.94 |
| | CLIP (standard) | 33.36 | 53.87 | 72.34 | 79.33 | 35.07 | 54.28 | 71.84 | 78.91 |
| | FAST-GOAL | **56.62** | **75.26** | **87.75** | **91.08** | **56.74** | **75.99** | **88.71** | **92.46** |

underperforming even standard CLIP extended with positional embedding interpolation to handle 248 tokens, which achieves 58.53% and 33.36% respectively. This demonstrates that processing the entire lengthy caption as a coherent sequence maintains contextual relationships across the full description more effectively than fragmenting it into independent sentences. However, simply extending token length through positional interpolation remains insufficient for truly understanding detailed descriptions, as evidenced by FAST-GOAL's substantial improvements to 74.27% on DOCCI and 56.62% on DCI.

These results highlight why explicit global-local alignment is essential. Long captions contain rich information about multiple objects, fine-grained attributes, spatial relationships, and contextual details that must be understood both holistically and locally. While positional interpolation enables processing lengthy text, it does not provide the supervision necessary to effectively leverage the detailed information within these descriptions. Our global-local alignment strategy addresses this through dual-level supervision: global contrastive learning ensures overall scene understanding by processing the entire caption coherently, while local contrastive learning with TSL enables fine-grained region-text associations. The substantial performance improvements of FAST-GOAL over both sentence averaging and extended CLIP validate that explicit fine-tuning with global-local alignment is essential for effective long-text understanding.

**Algorithm 1** Extract and pool patch tokens from bounding box coordinates

**Require:** Patch tokens $\mathbf{P} \in \mathbb{R}^{N \times d}$, bounding box coordinates $(x_1, y_1, x_2, y_2)$, patch size $p$, number of patches per row $n_p$
**Ensure:** Pooled feature vector $\mathbf{v} \in \mathbb{R}^d$
**// Convert bbox coordinates to patch indices**
$\text{patch}_{x_1}, \text{patch}_{y_1} \leftarrow \lfloor x_1/p \rfloor, \lfloor y_1/p \rfloor$
$\text{patch}_{x_2}, \text{patch}_{y_2} \leftarrow \lceil x_2/p \rceil, \lceil y_2/p \rceil$
**// Collect all patches overlapping with bbox**
$\mathcal{I} \leftarrow \{i \cdot n_p + j \mid \text{patch}_{y_1} \leq i < \text{patch}_{y_2}, \text{patch}_{x_1} \leq j < \text{patch}_{x_2}\}$
**// Average pooling over selected patches**
$\mathbf{v} \leftarrow \frac{1}{|\mathcal{I}|} \sum_{k \in \mathcal{I}} \mathbf{P}[k]$
**return** $\mathbf{v}$

TABLE XVI
PATCH-LEVEL LOCALIZATION EVALUATION ON DOCCI DATASET USING VIT-B/16 BACKBONE. GIVEN LOCAL TEXT DESCRIPTIONS, MODELS PREDICT CORRESPONDING IMAGE REGIONS BY SELECTING PATCHES WITH HIGHEST SIMILARITY. THE BEST PERFORMANCE IS SHOWN IN **BOLD**.

| Method | Top-5 (%) | Top-10 (%) | Top-15 (%) |
|---|---|---|---|
| CLIP | 43.25 | 57.78 | 67.57 |
| FAST-GOAL | **46.39** | **60.85** | **69.70** |

**Patch-Level Localization Evaluation.** To evaluate whether our method enhances region-level representations beyond image-level retrieval performance, we conduct patch-level localization experiments on the DOCCI dataset. Table XVI presents the evaluation setup and results. For each region-description pair in DOCCI where a specific image region corresponds to a descriptive sentence, we compute cosine similarity between the sentence embedding and all patch embeddings extracted from the entire image by the CLIP vision encoder. We then identify the patch with the highest similarity and check whether its center coordinate falls within the bounding box specifying the image region location. If the predicted patch location lies within this bounding box, we consider it a correct localization.

FAST-GOAL consistently outperforms CLIP across all evaluation thresholds. At Top-5, FAST-GOAL achieves 46.39% compared to CLIP's 43.25%. The improvements extend to higher thresholds, with Top-10 showing 60.85% compared to 57.78%, and Top-15 achieving 69.70% compared to 67.57%. These results validate that our global-local alignment training with TSL effectively enhances the model's ability to establish precise correspondences between textual descriptions and specific image regions. The consistent improvements across different thresholds confirm that FAST-GOAL not only improves image-level retrieval performance but also strengthens region-level representations, enabling better localization of objects and concepts described in text.

### C. Implementation Details for Patch Token Selection

In our Token Similarity-based Learning (TSL) method described in Eq. (9) of the main paper, we aggregate patch tokens from specific image regions based on bounding box

**Prompt**

"Generate ONE FALSE statement that is completely UNRELATED to this image. The statement should:
- Describe objects, scenes, or situations that have NO connection to what's actually shown in the image
- Sound natural and plausible as a description
- Be a complete sentence
- Be specific and detailed enough to blend with real descriptions
- NOT use phrases like "there is no", "doesn't show", "not visible" - instead describe something false as if it were true
- Be completely random and unrelated to the image content (e.g., if image shows outdoor scene, describe indoor objects; if image shows animals, describe vehicles, etc.)

Examples of good unrelated false statements:
- "A vintage typewriter sits on the mahogany desk."
- "The chef is garnishing the pasta with fresh basil leaves."
- "Two golden retrievers are playing fetch in the park."
- "The bookshelf is filled with leather-bound classical novels."

Generate ONLY the false unrelated statement, nothing else."

**Image**

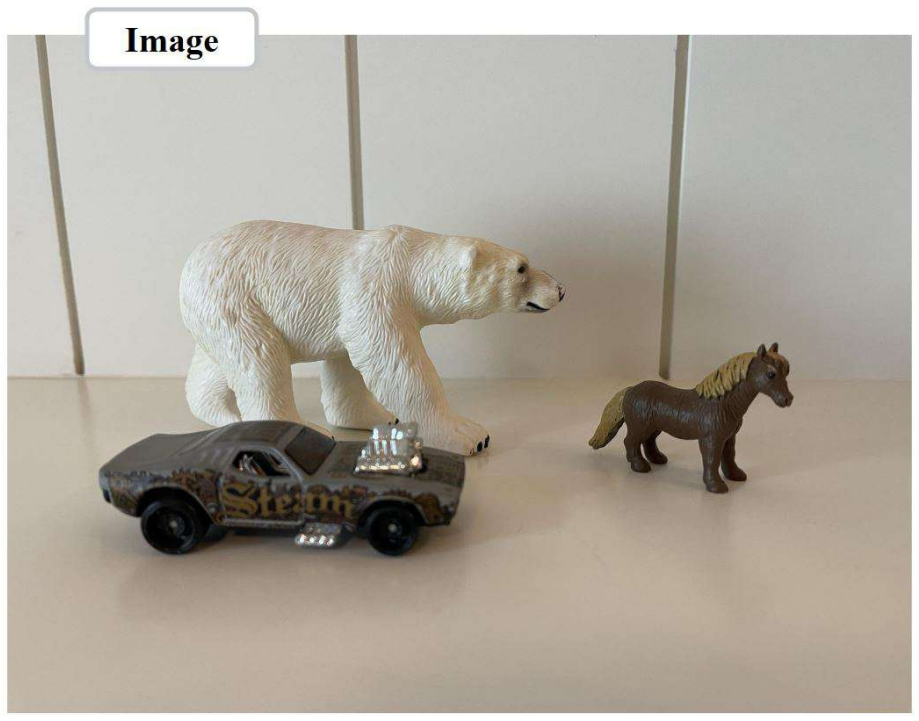

**Wrong descriptions in the lengthy caption**

An indoor close-up shot of a grey Hot Wheels roger dodger steam punk car with its silver engine components protruding from the hood facing the right in front of a toy polar bear and a small brown horse toy. A group of people are enjoying a picnic on the beach, with the sound of waves crashing in the background. The polar bear and horse are also facing the right. Polar bears positioned on all fours with its right side legs forward as if it's a walking. The black tip nose of the polar bear is slightly worn. The small brown horse stands with all four hooves flat Multiple Shadows of the objects are cast towards the left and right, indicating multiple light sources…

Fig. 8. Example of robustness evaluation with incorrect descriptions. False statements completely unrelated to the actual image content (shown in red) are generated using LLaVA-Next and inserted into the middle of original lengthy captions. The remaining portions contain correct descriptions of the image.

TABLE XVII
ROBUSTNESS ANALYSIS ON DOCCI DATASET WITH INCORRECT DESCRIPTIONS INJECTED INTO LENGTHY CAPTIONS USING VIT-B/16 BACKBONE. PERFORMANCE DROP INDICATES THE AVERAGE ABSOLUTE DECREASE IN R@1 ACROSS BOTH RETRIEVAL DIRECTIONS. FAST-GOAL DEMONSTRATES THE SMALLEST PERFORMANCE DEGRADATION. THE BEST PERFORMANCE IS SHOWN IN **BOLD**.

| Dataset | Method | Text to Image Recall@K | | | | Image to Text Recall@K | | | | Drop (%) |
|---|---|---|---|---|---|---|---|---|---|---|
| | | R@1 | R@5 | R@25 | R@50 | R@1 | R@5 | R@25 | R@50 | |
| DOCCI | CLIP (w/ Wrong) | 49.75 | 79.20 | 94.29 | 97.45 | 50.35 | 76.84 | 92.75 | 96.39 | -8.62 |
| | CLIP (Clean) | **58.53** | **85.57** | **97.16** | **98.82** | **58.80** | **85.18** | **96.73** | **98.45** | |
| | Long-CLIP (w/ Wrong) | 66.27 | 89.53 | 98.29 | 99.45 | 60.84 | 87.35 | 97.86 | 99.39 | -3.91 |
| | Long-CLIP (Clean) | **71.63** | **92.16** | **98.90** | **99.73** | **63.29** | **88.80** | **98.39** | **99.45** | |
| | FAST-GOAL (w/ Wrong) | 71.47 | 91.92 | 98.92 | 99.59 | 71.69 | 93.10 | 99.16 | 99.73 | **-2.89** |
| | FAST-GOAL (Clean) | **74.27** | **93.76** | **99.27** | **99.71** | **74.67** | **94.71** | **99.24** | **99.75** | |

coordinates obtained from FLISM. This section provides detailed implementation of the patch token selection process, particularly addressing cases where bounding boxes may be smaller than individual ViT [6] patch dimensions.

Algorithm 1 presents the complete procedure for extracting and pooling patch tokens from a given bounding box. The key design principle is to collect all patches that overlap with the bounding box region, ensuring comprehensive representation regardless of bbox size relative to patch dimensions. Specifically, we convert bounding box coordinates to patch indices using floor operations for the top-left corner and ceiling operations for the bottom-right corner. This conversion guarantees that all patches with any spatial overlap with the bounding box are included in the selection. For bounding boxes smaller than a single ViT patch, this approach ensures at least one patch token is selected, as the floor and ceiling operations will capture the patch containing the bbox region. For larger bounding boxes spanning multiple patches, all overlapping patches are collected to provide complete spatial coverage. The selected patch tokens are then aggregated through average pooling to obtain a unified feature representation $P_m$ that captures comprehensive information from the specified region. This robust extraction method enables effective learning of fine-grained correspondences between image regions and textual descriptions across varying object scales and spatial extents.

### D. Robustness Analysis with Incorrect Descriptions

To evaluate the robustness of our method when handling captions containing incorrect information, we conduct systematic experiments by intentionally injecting false statements

into the lengthy captions of the DOCCI [2] test set. Figure 8 illustrates the experimental setup. Using LLaVA-Next [15] with a carefully designed prompt, we generate completely unrelated false statements that describe objects, scenes, or situations not present in the actual images. These incorrect descriptions are inserted naturally into the middle of the original captions while preserving all correct descriptions, creating a realistic scenario where captions contain partial errors.

Table XVII presents the performance comparison across different methods when processing captions with injected incorrect descriptions. FAST-GOAL demonstrates the smallest performance degradation, experiencing only a 2.89% average drop in R@1 metrics compared to clean captions. In contrast, Long-CLIP [8] shows a 3.91% drop, while CLIP [5] exhibits a substantial 8.62% degradation. These results indicate that FAST-GOAL maintains more stable retrieval performance in the presence of partial caption errors.

This robustness can be attributed to our global-local alignment strategy. Since the majority of each caption still contains correct descriptions of the image, and our method learns to establish correspondences at both global and local levels through TSL, FAST-GOAL can effectively leverage the correct portions of the caption. The model's ability to maintain strong performance despite partial errors demonstrates that our global-local alignment enables focusing on the most reliable correspondences while maintaining comprehensive understanding of the correct portions in lengthy captions. This natural resilience to isolated incorrect statements validates the effectiveness of our training approach in learning robust multi-modal representations that can handle realistic scenarios where captions may contain occasional inaccuracies.

## References


[1] H. Choi, Y. K. Jang, and C. Eom, "GOAL: Global-local object alignment learning," in *Proceedings of the IEEE/CVF Conference on Computer Vision and Pattern Recognition*, 2025, pp. 4070–4079.

[2] Y. Onoe, S. Rane, Z. Berger, Y. Bitton, J. Cho, R. Garg, A. Ku, Z. Parekh, J. Pont-Tuset, G. Tanzer *et al.*, "DOCCI: Descriptions of connected and contrasting images," in *Proceedings of the European Conference on Computer Vision*, 2024, pp. 291–309.

[3] A. Kirillov, E. Mintun, N. Ravi, H. Mao, C. Rolland, L. Gustafson, T. Xiao, S. Whitehead, A. C. Berg, W.-Y. Lo *et al.*, "Segment anything," in *Proceedings of the IEEE/CVF International Conference on Computer Vision*, 2023, pp. 4015–4026.

[4] Y. Fang, B. Liao, X. Wang, J. Fang, J. Qi, R. Wu, J. Niu, and W. Liu, "You only look at one sequence: Rethinking transformer in vision through object detection," in *Advances in Neural Information Processing Systems*, 2021, pp. 26 183–26 197.

[5] A. Radford, J. W. Kim, C. Hallacy, A. Ramesh, G. Goh, S. Agarwal, G. Sastry, A. Askell, P. Mishkin, J. Clark *et al.*, "Learning transferable visual representations from natural language supervision," in *Proceedings of the International Conference on Machine Learning*, 2021, pp. 8748–8763.

[6] A. Dosovitskiy, L. Beyer, A. Kolesnikov, D. Weissenborn, X. Zhai, T. Unterthiner, M. Dehghani, M. Minderer, G. Heigold, S. Gelly *et al.*, "An image is worth 16x16 words: Transformers for image recognition at scale," in *Proceedings of the International Conference on Learning Representations*, 2021.

[7] J. Urbanek, F. Bordes, P. Astolfi, M. Williamson, V. Sharma, and A. Romero-Soriano, "A picture is worth more than 77 text tokens: Evaluating CLIP-style models on dense captions," in *Proceedings of the IEEE/CVF Conference on Computer Vision and Pattern Recognition*, 2024, pp. 26 700–26 709.

[8] B. Zhang, P. Zhang, X. Dong, Y. Zang, and J. Wang, "Long-CLIP: Unlocking the long-text capability of CLIP," in *Proceedings of the European Conference on Computer Vision*, 2025, pp. 310–325.

[9] T.-Y. Lin, M. Maire, S. Belongie, J. Hays, P. Perona, D. Ramanan, P. Dollár, and C. L. Zitnick, "Microsoft COCO: Common objects in context," in *Proceedings of the European Conference on Computer Vision*, 2014, pp. 740–755.

[10] P. Young, A. Lai, M. Hodosh, and J. Hockenmaier, "From image descriptions to visual denotations: New similarity metrics for semantic inference over event descriptions," in *Transactions of the Association for Computational Linguistics*, 2014, pp. 67–78.

[11] Q. Sun, Y. Fang, L. Wu, X. Wang, and Y. Cao, "EVA-CLIP: Improved training techniques for CLIP at scale," in *arXiv preprint arXiv:2303.15389*, 2023.

[12] M. Tschannen, A. Gritsenko, X. Wang, M. F. Naeem, I. Alabdulmohsin, N. Parthasarathy, T. Evans, L. Beyer, Y. Xia, B. Mustafa *et al.*, "Siglip 2: Multilingual vision-language encoders with improved semantic understanding, localization, and dense features," 2025.

[13] F. Faghri, P. K. A. Vasu, C. Koc, V. Shankar, A. Toshev, O. Tuzel, and H. Pouransari, "Mobileclip2: Improving multi-modal reinforced training," 2025.

[14] K. Wu, H. Peng, Z. Zhou, B. Xiao, M. Liu, L. Yuan, H. Xuan, M. Valenzuela, X. S. Chen, X. Wang *et al.*, "Tinyclip: Clip distillation via affinity mimicking and weight inheritance," in *Proceedings of the IEEE/CVF International Conference on Computer Vision*, 2023, pp. 21 970–21 980.

[15] H. Liu, C. Li, Y. Li, B. Li, Y. Zhang, S. Shen, and Y. J. Lee, "LLaVA-NeXT: Improved reasoning, OCR, and world knowledge," 2024.


\end{document}